\documentclass{ecai} 

\usepackage{latexsym}
\usepackage{amssymb}
\usepackage{amsmath}
\usepackage{amsthm}
\usepackage{booktabs}
\usepackage{enumitem}
\usepackage{graphicx}
\usepackage{color}

\usepackage{times}
\usepackage{soul}
\usepackage{url}
\usepackage[hidelinks]{hyperref}
\usepackage[utf8]{inputenc}
\usepackage[small]{caption}
\usepackage{algorithm}
\usepackage{algorithmic}
\usepackage[switch]{lineno}
\usepackage{multirow}
\usepackage{indentfirst}
\usepackage{makecell}
\usepackage{subfigure}
\usepackage{float}
\graphicspath{{images}}
\usepackage{stfloats}
\usepackage{indentfirst}
\usepackage{colortbl}
\usepackage{arydshln}
\usepackage{CJKutf8}
\usepackage{listings}
\usepackage{xcolor}
\definecolor{grey}{rgb}{0.8, 0.8, 0.8}

\newtheorem{definition}{Definition}

\newcommand{\BibTeX}{B\kern-.05em{\sc i\kern-.025em b}\kern-.08em\TeX}

\begin{document}


\begin{frontmatter}


\paperid{7407} 


\title{OMoE: Diversifying Mixture of Low-Rank Adaptation by Orthogonal Finetuning}


\author[A,B,C]{\fnms{Jinyuan}~\snm{Feng}}
\author[A,B,C]{\fnms{Zhiqiang}~\snm{Pu}\thanks{Corresponding Author.}}
\author[A,B,C]{\fnms{Tianyi}~\snm{Hu}} 
\author[A,B,C]{\fnms{Dongmin}~\snm{Li}}
\author[A,B,C]{\fnms{Xiaolin}~\snm{Ai}} 
\author[D]{\fnms{Huimu}~\snm{Wang}} 

\address[A]{Institute of Automation, Chinese Academy of Sciences}
\address[B]{The Key Laboratory of Cognition and Decision Intelligence for Complex Systems, Institute of Automation, Chinese Academy of Sciences}
\address[C]{School of Artificial Intelligence, University of Chinese Academy of Sciences}
\address[D]{JD.COM}


\begin{abstract}
Building mixture-of-experts (MoE) architecture for low-rank adaptation (LoRA) is emerging as a potential direction in parameter-efficient fine-tuning (PEFT) for its modular design and remarkable performance. However, simply stacking the number of experts cannot guarantee significant improvement. In this work, we first conduct qualitative analysis to indicate that experts collapse to similar representations in vanilla MoE, limiting the capacity of modular design and computational efficiency. Ulteriorly, our analysis reveals that the performance of previous MoE variants may be limited by a lack of diversity among experts. Motivated by these findings, we propose Orthogonal Mixture-of-Experts (OMoE), a resource-efficient MoE variant  that trains experts in an orthogonal manner to promote diversity. In OMoE, a Gram-Schmidt process is leveraged to enforce that the experts' representations lie within the Stiefel manifold. By applying orthogonal constraints directly to the architecture, OMoE keeps the learning objective unchanged, without compromising optimality. Our method is simple and alleviates memory bottlenecks, as it incurs minimal experts compared to vanilla MoE models. Experiments on diverse commonsense reasoning benchmarks demonstrate that OMoE can consistently achieve stable and efficient performance improvement when compared with the state-of-the-art methods while significantly reducing the number of required experts.
\end{abstract}

\end{frontmatter}


\section{Introduction}
Large language models (LLMs) offer a versatile and task-agnostic foundation that underpins an extensive array of applications, including text generation~\cite{zhuang2023toolqa,song2023label}, question answering~\cite{huang2024c}, reasoning~\cite{wang2024ts}, and personalized chat-bots~\cite{achiam2023gpt,chan2023gpt}. Due to the computational and memory resources required by full-parameter fine-tuning, parameter-efficient fine-tuning (PEFT) methods have emerged~\cite{hu2021lora,hayou2024loraplus,liu2024dora,kalajdzievski2023rank,li2021prefix}. As a popular PEFT method, low-rank adaptation (LoRA) is characterized by freezing the backbone parameters while lightweight low-rank parameter matrices are fine-tuned. LoRA marks a significant advancement in the practical use of LLMs.
\begin{figure*}[t]
    \centering
    \includegraphics[width=0.85\textwidth]{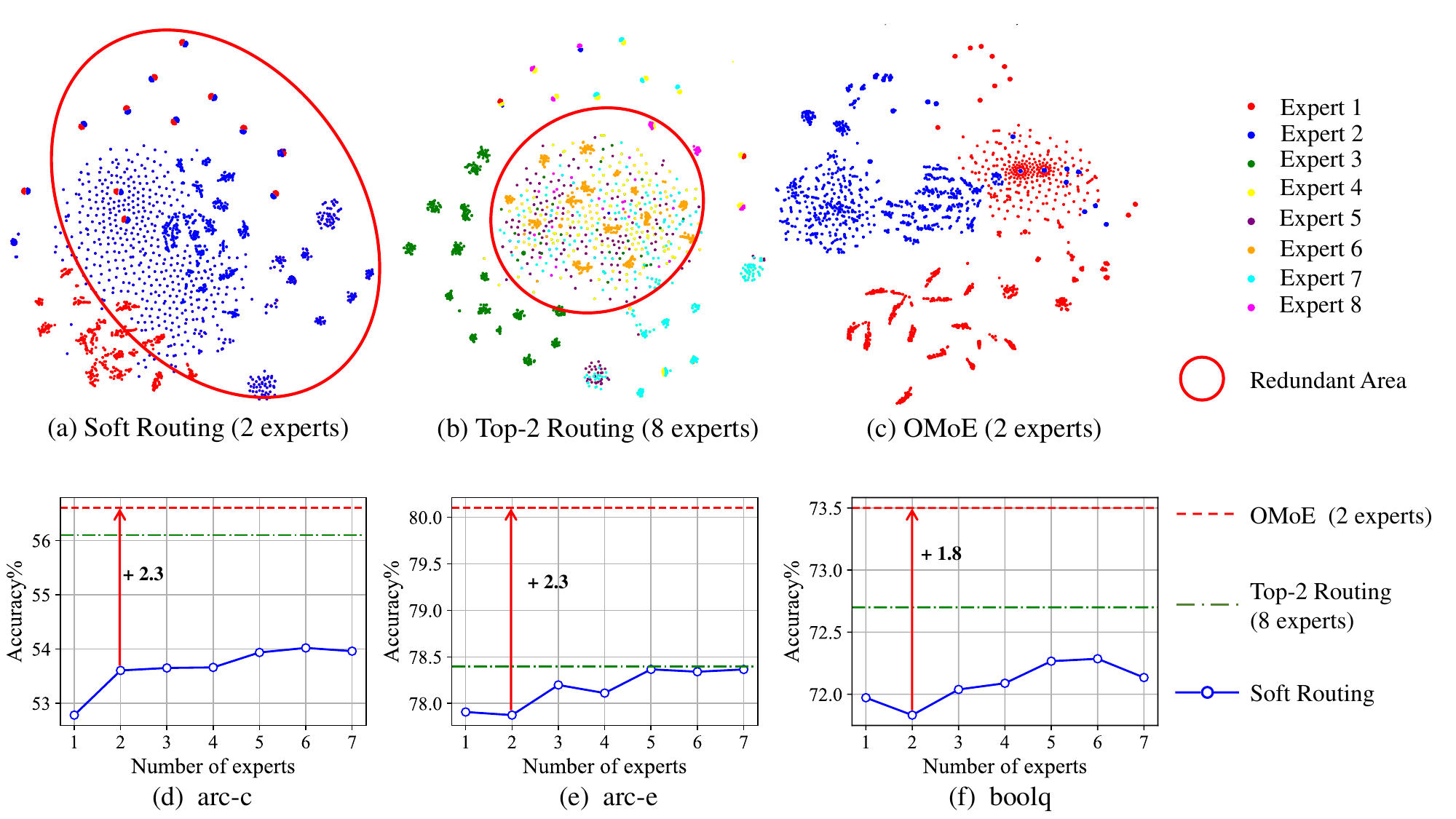}
    \caption{(a), (b) and (c) visualize representations of two activated experts using soft routing, Top-2 routing and our OMoE method, while Top-2 routing utilizes a total of 8 experts. The representations are obtained from layer 8 of LLaMA-2 7B by projecting the vectors into 2D space via t-SNE for qualitative analysis. The visualization of more layers is provided in Appendix~\ref{differentmethod}. (d), (e) and (f) illustrate performance comparison between soft routing, Top-2 routing and OMoE.}
\label{motivation_example}
\vspace{8pt}
\end{figure*}

Despite LoRA's effectiveness, the issue of compromised quality in heterogeneous tasks motivates recent studies to explore the combination of mixture-of-experts (MoE) and LoRA for its modular design and versatility. MoE enables flexible and potentially more generalizable handling of diverse data within a single model by routing them to specialized experts~\cite{shi2024unchosen,li2024mixlora,liu2023moelora,zhang2024milora,tian2024hydralora,lin2024teamlora,chen2024llava}. For more details, we provide a section for related works in Section~\ref{relatedworks}. In principle, for each input token, activated experts need to act synergistically to generate the output, which presents a compelling research question: \textit{How can activated experts collaborate efficiently to maximize the utilization of the model's capacity?}

In practice, MoE stacks more experts and parameters to tackle complex problems, as it enables a broader spectrum of specialized knowledge. Recent works~\cite{qian2024scaling,shi2024unchosen} indicate that merely raising the number of experts does not lead to significant improvements, and models' performance approximately follows a logistic growth pattern. A direct hypothesis is that most experts collapse into similar distributions, thus lacking diversity and contributing minimally to overall performance. Our visualization of representation distributions in MoE under various routing strategies verifies this hypothesis. As shown in Fig.\ref{motivation_example}(a)(b), soft routing and top-2 routing in vanilla MoE exhibit apparent redundant areas in the representation space, indicating that experts redundantly learn the same feature, which leads to the underutilization of MoE's modular design.

MoE, similar to deep ensemble learning, has widely benefited from diversity to enhance robustness and efficiency~\cite{kim2024sample}. However, existing MoE-based methods do not effectively address the issue of diversity, which is essential for enhancing models' efficiency. Although selective updates in top-k routing promote dissimilarities among experts, it is not sufficient for significantly improving the diversity of experts. Current solutions promote diversity by introducing an additional objective as diversity booster, which alter the main learning objective~\cite{lan2024contrastive,dou2023loramoe}. In contrast, the novelty of OMoE lies in its hard constraint instead of introducing regularization loss. OMoE applies orthogonal constraints directly without compromising optimality, achieving diversity effectively. A qualitative illustration of OMoE is presented in Fig.~\ref{motivation_example}(c).

Consequently, we propose a novel method named Orthogonal Mixture-of-Experts (OMoE), which is a resource-efficient MoE variant, to promote the diversity of experts. Specifically, OMoE orthogonalizes the representations generated by a mixture of experts via Gram-Schmidt process, thus favoring dissimilarity and preserving the hyperspherical energy. The Gram-Schmidt process applies orthogonal constraints directly to the architecture, thus OMoE keeps the learning objective unchanged, without compromising optimality. Mathematically, the manipulated orthogonal representations belong to particular Riemannian manifold where the inner product is defined, known as Stiefel manifold~\cite{li2020efficient}. In this paper, we focus on fine-tuning MoE models to enhance diversity and improve the performance of LLMs. Our extensive experiments and analyses across diverse benchmarks demonstrate that OMoE consistently outperforms state-of-the-art PEFT baselines while significantly reducing tunable parameters by approximately 75\%. Summary of our contributions:
\begin{itemize}
\item We systematically analyze that most experts collapse into similar distributions in MoE from the perspective of representations. And we show that promoting diversity is essential for achieving efficient PEFT.
\item We propose a novel and resource-efficient MoE variant, named OMoE, that trains experts in an orthogonal manner to effectively promote diversity.
\item Extensive experiments demonstrate that OMoE can consistently achieve stable performance while significantly reducing tunable parameters.
\end{itemize}


\section{Preliminaries and Analysis}
\subsection{LoRA Basics}
LoRA~\cite{hu2021lora} achieves performance comparable to fine-tuning on multiple benchmarks by introducing low-rank matrices into the model’s architecture, facilitating effective adaptation to specific tasks while keeping the original model weights $W_{0}$ frozen. A LoRA block consists of two sequential low-rank matrices $A$ and $B$ to fit the residual weights for fine-tuning. The updated forward computation is calculated through:

\begin{equation}
    y\prime=y + \Delta y =W_{0}x + BAx   
    \label{eq:lora}
\end{equation}
, where $y' \in \mathbb{R}^d$ is the output, $y$ is the original model's output and $x \in \mathbb{R}^k$ denotes the input token. $B \in \mathbb{R}^{d×r}, A \in \mathbb{R}^{r×k}$ with rank $r \ll \min(d, k)$. Normally, matrix $B$ is initialized with zeroes and matrix $A$ is initialized with Kaiming Uniform~\cite{he2015delving} to force $\Delta y = 0$. 
\subsection{LoRA's MoE Variants}
MoE variants involve replacing the original LoRA modules with $n$ parallel minor LoRA submodules, referred to as experts, denoted by $\{E_i=B_iA_i\}_{i=1}^n$. During training and inference, a router $g(\cdot;\mathbf{G})$, parameterized by $\mathbf{G}$, assigns the input token $x$ to a subset of LoRA experts, $g(\cdot;\mathbf{G}) \in \mathbb{R}^n$. Typically, the router $g$ consists of a simple linear layer followed by a softmax to normalize coefficients and then the forward computation is modified as:
\begin{equation}
    y\prime=y + \Delta y =W_{0}x + \sum_{i=1}^ng_i(x;\mathbf{G})E_i(x).
    \label{eq:moelora}
\end{equation}
Here, $g_i(x;\mathbf{G})$ indicates the router’s output for the $i$-th expert, and $g(x;\mathbf{G})$ can be designed as top-k routing, top-p routing, soft routing, or top-1 switch routing.

\subsection{MoE's Dilemma: An Exploratory Analysis}
Compared with full fine-tuning (FFT), PEFT exhibits a notable performance disparity with limited trainable parameters, particularly in diverse or heterogeneous training corpus. This disparity can be alleviated by MoE technique. However, as depicted in Fig.~1(d)(e)(f), simply increasing the number of experts does not guarantee improved performance in vanilla MoE. Inspired by this finding, we investigate the performance of MoE using different routing strategies, specifically comparing soft routing and top-2 routing, detailed as follows:

\textbf{Soft routing.} Soft routing in MoE allows for a weighted average of inputs to be assigned to multiple experts, enabling more flexible and adaptive processing. This means $g(x;\mathbf{G})_i$ is nonzero for all experts.

\textbf{Top-k routing.} Top-k routing is a well-performing routing strategy of LLM and PEFT, which activates the $k$ experts with the highest values in $g(x;\mathbf{G})$. Then, the renormalized router value for the $i$-th expert, $\hat{g}(x;\mathbf{G})_i$, is defined as follows:
\begin{equation}
\hat{g}_i(x)=
\begin{cases}
\frac{g_i(x)}{\sum_{j\in\operatorname{top}(\mathbf{w},2)}g_j(x)}, &i\in\operatorname{top}(g(x),k) \\
0, &i\notin\operatorname{top}(g(x),k),
\end{cases}
\end{equation}
where $\operatorname{top}(g(x),k)$ returns the indices of the largest $k$ elements in $g(x)$.

From the perspective of model performance, as illustrated in Fig.\ref{motivation_example}(c)(d)(e), top-2 routing exhibits superior performance compared to soft routing. In terms of representation, we visualized representations of two activated experts using soft routing and top-2 routing, with the latter utilizing a total of eight experts. A qualitative illustration is presented in Fig. \ref{motivation_example}(a)(b). The representations in top-2 routing show greater diversity compared to soft routing, suggesting the benefits of diversity. However, redundant areas stick in top-2 routing. Our analysis supports the conclusions of \cite{chi2022representation} that if the hidden states are routed to activated experts, they will be pushed closer together.
In this paper, we would like to force the representations to be more diverse, so that they can be more expressive and discriminative. A comprehensive analysis of Fig.~\ref{motivation_example}, leads us to identify three key observations:

\textbf{Observation 1:} \textit{Deploying multiple LoRA experts does not yield significant performance improvements. From a representation perspective, substantial representation redundancy becomes a bottleneck for model performance.}

\textbf{Observation 2:} \textit{Top-2 routing demonstrates superior performance compared to soft routing, which can be attributed to its greater representational diversity. Enhancing the diversity of experts is essential for optimizing model performance.}

\textbf{Observation 3:} \textit{Top-2 routing enhances the diversity of experts by selectively updating them, which inevitably introduces a large number of experts. The issue of representation redundancy among these experts persists in top-2 routing.}

Building upon our observations, OMoE is proposed to enhance expert diversity and reduce representation redundancy. In our method, experts are trained orthogonally, with their representations residing within the Stiefel manifold. As illustrated in Fig.~\ref{motivation_example} (c), OMoE effectively reduces representation redundancy while significantly reducing tunable parameters. OMoE is trained end-to-end based on MoE and LoRA, which will be detailed in Section~\ref{OMoE:method}. This method facilitates flexibility and applicability, even in diverse or heterogeneous training corpus.


\section{OMoE: Orthogonal Mixture-of-Experts}

\label{OMoE:method}
\begin{figure}[ht]
    \centering
    \includegraphics[width=0.45\textwidth]{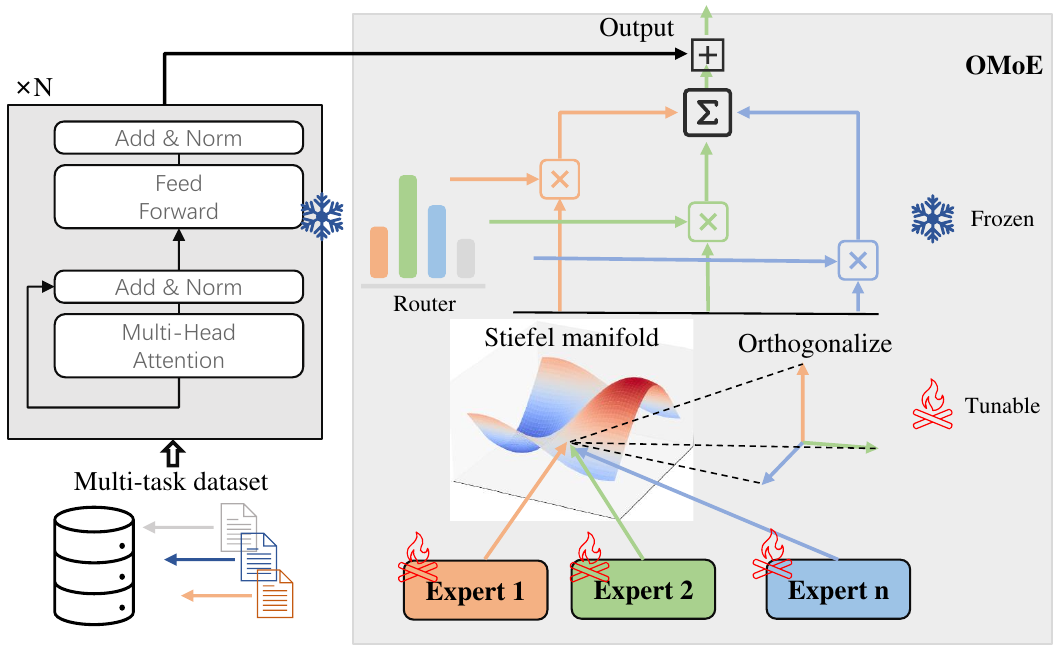}
    \caption{The architecture of OMoE. During the fine-tuning stage, expert representations are constrained to lie within the Stiefel manifold, thereby achieving orthogonalization.}
\label{OMoE}
\vspace{15pt}
\end{figure}

In this section, we first introduce the basics of the Stiefel manifold, and then elaborate on the architectural design of OMoE, as illustrated in Fig.~\ref{OMoE}.

\subsection{Optimization within Stiefel Manifold}
The Stiefel manifold, as a smooth and differentiable Riemannian manifold, has been widely used in neural network training~\cite{qiu2023controlling,hendawy2023multi,liu2023diversifying}. In our context, we study Riemannian optimization for MoE in PEFT, which ensures that expert representations satisfy orthonormality constraints. A set of $k$ expert representations in $\mathbb{R}^{d}$ for a given token in LoRA can be transformed into orthonormal representations $E=[e_{1},\ldots,e_{k}]\in\mathbb{R}^{d\times k}$ using the Gram-Schmidt process or Cayley transformation. These orthonormal representations belong to a topological space known as the Stiefel manifold, making MoE-based PEFT equivalent to Riemannian optimization on the Stiefel manifold.
\begin{definition}(Riemannian Manifold)
A Riemannian manifold ($\mathcal{M}, \rho$) is a smooth manifold $\mathcal{M}$ equipped with a Riemannian metric $\rho$ defined as the inner product on the tangent space $T_{x}\mathcal{M}$ for each point $x$,  $\rho_{x}(\cdot,\cdot): T_{x}\mathcal{M} \times T_{x}\mathcal{M} \to \mathbb{R}$.
\end{definition}
The Stiefel manifold $S(d,k)$ (where $d\geq k$) is a Riemannian manifold consisting of all $d\times k$ orthonormal matrices $S\in \mathbb{R}^{d\times k}$ such that $S^TS=I$. The Stiefel manifold $S(d,k)$ can be regarded as an embedded submanifold of a Euclidean space. Consequently, the Riemannian metric $\rho$ is defined using the Euclidean metric as: $\rho_E(Z_1,Z_2)=tr(Z_1^TZ_2)$, where $Z_1,Z_2$ are vectors in the tangent $T_S\mathcal{M}$.

We assume that MoE consists of a mixture of experts with learnable parameters $\phi=[\phi_1,...,\phi_k]$, and their representations are formulated as $E_\phi=[e_{1},\ldots,e_{k}]$. To ensure that the expert representations lie within the Stiefel manifold and promote diversity, we impose an orthogonality constraint on the optimization target of supervised fine-tuning (SFT)~\cite{ouyang2022training}. This leads to the following formulation:
\begin{equation}
    \begin{aligned}
 & \max_{\phi} & & J(\phi) \\
 & \mathrm{s.t.} & & E_\phi^TE_\phi=I_k,
\end{aligned}
\label{orth}
\end{equation}
where $I_k\in\mathbb{R}^{k\times k}$ is the identity matrix and $J(\phi)$ is the objective function for SFT. During the training process of $J(\phi)$, we ensure that the orthogonality constraint is satisfied by applying the Gram-Schmidt process to enforce orthogonalization. 
\begin{table*}[b]
    \centering
\resizebox{0.88\textwidth}{!}{
    \begin{tabular}{lcccccccccc}
        \toprule
        \textbf{Model} & \textbf{Method} & \textbf{Params} & \textbf{ARC-e} & \textbf{ARC-c} & \textbf{BoolQ} & \textbf{OBQA} & \textbf{PIQA} & \textbf{SIQA} & \textbf{HellaS} & \textbf{WinoG} \\ 
        \midrule
        \multirow{5}{*}{LLaMA-2 7B}
          & LoRA & 2.9\% & 73.8  & 50.9  & 62.2  & 80.4  & 82.1  & 69.9 & 88.4 & 66.8  \\
        ~ & DoRA & 2.9\% & 76.5  & \textbf{59.8}  & 71.7  & 80.6  & 82.7  & 74.1 & 89.6 & 67.3  \\ 
        ~ & MixLoRA & 2.9\% & 78.4* & 56.1* & 72.7 & \textbf{81.6} & 83.2 & 78.0 & 92.8* & \textbf{76.8}  \\ 
        ~ & \textbf{OMoE-LoRA} & \colorbox{grey}{0.73\%} & \underline{79.3} & \underline{56.6} & \textbf{73.5} & \underline{80.6} & \textbf{84.5} & \textbf{79.6} & \underline{92.9} & 67.2 \\
        ~ & \textbf{OMoE-DoRA} & \colorbox{grey}{0.73\%} & \textbf{80.1} & 55.3 & \underline{73.1} & 80.0 & \underline{83.3} & \underline{79.0} & \textbf{93.0} & \underline{68.4} \\ 
        \midrule
        \multirow{5}{*}{LLaMA-3 8B} & LoRA & 2.6\% & 89.0  & 75.7  & 67.2  & 85.0  & 80.7  & 78.3 & 74.2 & 75.3 \\
        ~ & DoRA & 2.6\% & 88.1  & 76.4  & 61.7  & 80.6  & 82.3  & 76.2 & 78.8 & 83.7 \\ 
        ~ & MixLoRA  & 3.0\% & 86.5 & \textbf{79.9} & \textbf{75.0} & 84.8 & 87.6 & 78.8 & 93.3 & 82.1  \\ 
        ~ & \textbf{OMoE-LoRA} & \colorbox{grey}{0.75\%} & \textbf{89.1} & \underline{76.7} & 73.9 & \textbf{86.6} & \textbf{88.4} & \underline{79.7} & \textbf{95.1} & \underline{84.8}  \\ 
        ~ & \textbf{OMoE-DoRA} & \colorbox{grey}{0.75\%} & \underline{89.0} & 76.7 & \underline{74.7} & \underline{86.4} & \underline{87.8} & \textbf{79.9} & \underline{95.0} & \textbf{85.2}  \\ 
        \midrule
        \multirow{5}{*}{LLaMA-2 13B-bf} & LoRA & 2.4\% & 82.7*  & 68.3*  & 72.0*  & \textbf{84.6*}  & 84.4*  & 80.0 & 94.3 & 81.9 \\ 
        ~ & DoRA & 2.4\% & 81.6*  & 65.6*  & 72.5*  & 81.2*  & 82.8*  & 80.1 & 92.5* & 82.4 \\ 
        ~ & MixLoRA & 2.5\% & \textbf{83.5}  & 67.5*  & \underline{75.3*}  & 82.8*  & 84.8*  & 79.6* & 94.5* & \textbf{84.5} \\ 
        ~ & \textbf{OMoE-LoRA} & \colorbox{grey}{0.63\%} & 82.8  & \underline{68.3}  & \textbf{75.3}  & \underline{84.2}  & \textbf{86.5}  & \underline{80.3} & \textbf{95.0} & 81.9 \\
        ~ & \textbf{OMoE-DoRA} & \colorbox{grey}{0.63\%} & \underline{82.9}  & \textbf{68.9}  & 75.2  & 84.0  & \underline{85.5}  & \textbf{80.6} & \underline{94.9} & \underline{82.5} \\
        \bottomrule
    \end{tabular}}
    \caption{The overall comparison of different PEFT methods for single-task learning, using base models with different numbers of parameters. Reported results are accuracy scores and most of the results are extracted from the original papers, and * indicates that the results are reproduced by running the provided source code. Bold and underline indicate the best and the second-best results.}
    \label{table:single_task_results}
    \vspace{4mm}
\end{table*}
\subsection{OMoE Forward}
The fundamental idea behind OMoE is to maximize the representation capacity of experts while avoiding the collapse of similar experts, thereby enhancing their diversity. In OMoE, to generate a response, the input token has to go through the transformer blocks in LLMs to obtain a hidden state. We denote the hidden state of the input token before Transformer layer $l$ as $h^l\in\mathbb{R}$. The forward computation of OMoE is formulated as:
\begin{equation}
    h^{l+1}= W_{0}h^l + \sum_{i=1}^ng_i(h^l)E^\prime_i(h^l),
\end{equation}
\begin{equation}
    E^\prime= GS(E),
\end{equation}
where $W_{0}$ represents the frozen pretrained weights of the feedforward network or attention layer. $g(h^l)$ denotes the router employed to assign coefficients to specific experts for different hidden states, and $E^\prime$ represents the orthogonalized expert representations obtained after applying the Gram-Schmidt process, denoted as $GS(\cdot)$. Specifically, in $GS(\cdot)$, given the representations of experts $E=[e_{1},\ldots,e_{k}]$, we use the Gram-Schmidt process to obtain a set of k-orthonormal representations of experts $E^\prime=[e_{1}^\prime,\ldots,e_{k}^\prime]$:
\begin{equation}
    e_k^\prime=e_k-\sum_{i=1}^{k-1}\frac{\langle e_i^\prime,e_k\rangle}{\langle  e_i^\prime, e_i^\prime\rangle} e_i^\prime.
\end{equation}
The Gram-Schmidt process maps a set of linearly independent vectors $E$ to a set of orthonormal vectors $E^\prime$. The representation of the $i$-th expert $e_i$ is projected in the orthogonal direction to the subspace spanned by the representation of all $i-1$ experts, which can satisfy the hard constraint in Eq.~\ref{orth}. Fig.~\ref{OMoE} presents an illustration of how OMoE works. In this figure, the representations of experts are constrained to lie within the Stiefel manifold, resulting in a distinct difference in the directions of the vectors.

\section{Experiments}
In this section, we conduct a series of experiments and offer a succinct interpretation of the results to evaluate OMoE.
\subsection{Experiment Setting}
\textbf{Datasets and Benchmarks.} We conduct experiments on a collection of tasks: (a) Diverse commonsense reasoning datasets, including common-sense question-answering tasks, ARC-e and ARC-c~\cite{clark2018think}, OpenBookQA~\cite{mihaylov2018can}, PIQA~\cite{bisk2020piqa}, SocialIQA~\cite{sap2019socialiqa}, BoolQ~\cite{clark2019boolq}. (b) A science completion task, Hellaswag~\cite{zellers2019hellaswag}. (c) A fill-in-the-blank task, Winogrande~\cite{sakaguchi2021winogrande}. We utilize the PEFT framework provided by ~\cite{hu2023llm,li2024mixlora} for training on these datasets. We choose LLaMA-2 7B, LLaMA-3 8B and LLaMA-2 13B as our backbone models. The detailed statistics and evaluation metrics can be found in Appendix~\ref{sec:appendix_datasets}.

\textbf{Baselines.} We compare our method with the popular and well-performing baselines. First, we compare OMoE with LoRA, its variants and MoE variants: 1) LoRA~\cite{hu2021lora}; 2) DoRA~\cite{liu2024dora}; 3) MixLoRA~\cite{li2024mixlora}; 4) AdaLoRA~\cite{zhang2023adalora}; 5) MoELoRA~\cite{liu2023moelora}; 6) MiLoRA~\cite{zhang2024milora}. Then, we compare it with other recent PEFT baselines: 1) Parallel-Adapter~\cite{he2021towards}; 2) Learned-Adapter~\cite{zhang2023learned}; 3) P-tuning v2~\cite{liu2021p}; 4) IAPT~\cite{zhu2024iapt}; 5) BitFit~\cite{zaken2021bitfit}; 6)$(\text{IA})^3$~\cite{liu2022few}; 7) SSP~\cite{hu2022sparse}. Most of the results are extracted from the original papers, and some baselines are reproduced by running the provided source code. In the main results subsection, we choose LoRA, DoRA, and MixLoRA as representative baselines and the comparison of other baselines is provided in Appendix~\ref{Results_other_baselines}.

\textbf{Settings.} To evaluate the effectiveness of our method, we apply it on the basis of LoRA and DoRA, which are labeled as OMoE-LoRA and OMoE-DoRA respectively in the experiment. Both OMoE-LoRA and OMoE-DoRA are configured with rank r = 16, incorporating 2 experts and a soft routing mechanism. When fine-tuning pretrained LLMs, we only consider the SFT setting. After receiving a prompt or instruction, all the predictions are generated using the language modeling output. Due to space limitations, other experimental settings for the baseline methods and the training procedure are in Appendix~\ref{sec:experimental settings}.
\begin{table*}[tb!]
\centering
\resizebox{0.88\textwidth}{!}{
\begin{tabular}{ccccccccc}
\toprule
{\textbf{Method}}   &   \textbf{Params}   &   {\textbf{ST/MT} }    &     \textbf{ARC-e}   &     \textbf{ARC-c}   &   \textbf{BoolQ}   &   \textbf{OBQA}  &  \textbf{PIQA}    &   {\textbf{Avg.}}  \\
\hline

\multirow{2}*{LoRA }  &     \multirow{2}*{2.9\% }   &  ST   &     73.8    &    50.9   &    62.2   &    80.4   &    82.1       &   69.9   \\    

&     &   MT     &      61.3 ( \textcolor{red}{-12.5} )    &  55.7 ( \textcolor{blue}{+4.8} )  &   66.7  ( \textcolor{blue}{+4.5} )  &  71.6  ( \textcolor{red}{-8.8} )    &   72.4   ( \textcolor{red}{-9.7} )     &   65.5 ( \textcolor{red}{-4.4} )      \\

\hdashline

\multirow{2}*{DoRA}   &    \multirow{2}*{2.9\%}   &   ST   &    76.5   &   59.8   &   71.7  &    80.6   &    82.7    &     74.3   \\

&     &   MT     &   64.5  ( \textcolor{red}{-12} )     &  54.1 ( \textcolor{red}{-5.7} )  &   65.4 ( \textcolor{red}{-6.3} )  &     75.8 ( \textcolor{red}{-4.8} )    &   71.9 ( \textcolor{red}{-10.8} )   &      66.3 ( \textcolor{red}{-8} )     \\

\hdashline

\multirow{2}*{MOELoRA}  &    \multirow{2}*{1.0\% }    &    ST    &      76.8   &  60.2   &     72.0     &    81.1    &     82.7    &    74.6    \\

&     &   MT     &    76.1   ( \textcolor{red}{-0.7} )     &  59.3   ( \textcolor{red}{-0.9} )      &   71.5 ( \textcolor{red}{+0.1} )   &  80.7 ( \textcolor{red}{-0.4} )      &   82.1  ( \textcolor{red}{-0.3} )      &    73.9   ( \textcolor{red}{-0.5} )   \\

\hdashline

\multirow{2}*{MiLoRA  }  &    \multirow{2}*{0.93\%}     &    ST   &   77.8   &  61.2   &     72.8   &      81.7     &  83.3    &    75.4       \\

&       &   MT     &   77.4 ( \textcolor{red}{-0.4} )    &  61.5 ( \textcolor{red}{+0.3} )   &  72.3 ( \textcolor{red}{-0.3} )   &  81.3 ( \textcolor{red}{-0.4} )    &  83.5 ( \textcolor{red}{+0.3} )    &       75.2  ( \textcolor{red}{-0.1} )      \\

\hdashline

\multirow{2}*{MixLoRA}  &    \multirow{2}*{2.9\% }    &    ST    &      78.4   &  56.1   &     72.7     &    81.6    &     83.2    &    74.4    \\

&     &   MT     &    76.6   ( \textcolor{red}{-1.8} )     &  64.2   ( \textcolor{blue}{+8.1} )      &   71.2 ( \textcolor{red}{-1.5} )   &  81.6 ( \textcolor{red}{-0.0} )      &   82.7  ( \textcolor{red}{-0.5} )      &    75.3   ( \textcolor{blue}{+0.9} )   \\

\hdashline

\multirow{2}*{OMoE-LoRA}  &    \multirow{2}*{\textbf{0.73\% }}    &    ST    &      79.3   &  56.6   &     73.5     &    80.6    &     84.5    &    74.9    \\

&     &   MT     &    79.8   ( \textcolor{blue}{+0.5} )     &  66.8   ( \textcolor{blue}{+10.2} )      &   72.4 ( \textcolor{red}{-1.1} )   &  76.8 ( \textcolor{red}{-3.8} )      &   81.6  ( \textcolor{red}{-2.9} )      &    75.5   ( \textcolor{blue}{+0.3} )   \\

\hdashline

\multirow{2}*{OMoE-DoRA}  &    \multirow{2}*{\textbf{0.73\%} }    &    ST    &      80.1   &  55.3   &     73.1     &    80.0    &     83.3    &    74.4    \\

&     &   MT     &    79.8   ( \textcolor{red}{-0.3} )     &  67.4   ( \textcolor{blue}{+12.1} )      &   70.6 ( \textcolor{red}{-2.5} )   &  78.0 ( \textcolor{red}{-2.0} )      &   82.9  ( \textcolor{red}{-0.4} )      &    \textbf{75.7}   ( \textcolor{blue}{+1.3} )   \\

\bottomrule
\end{tabular}}
\caption{\label{tab:results_main_multi_task} The overall comparison of different PEFT methods for multi-task learning. The backbone model is LLaMA-2 7B. ST refers to the single-task setup, while MT refers to the multi-task setup. Reported results are accuracy scores, with the differences between MT and ST indicated in red font for decreases and in blue font for increases. } 
\label{multi-task}
\vspace{8pt}
\end{table*}

\begin{table}[tb!]
\centering
{
\begin{tabular}{cccccc}
\toprule
{\textbf{Method}}   &   \textbf{ARC-e}   &  \textbf{ARC-c}   & \textbf{BoolQ}   & {\textbf{OBQA} } & {\textbf{Avg.} }      \\
\hline

MixLoRA  &     84.8   &  75.0 & 75.0 & 85.2 & 80.0   \\    
\hdashline
OMoE-LoRA  &    \textbf{85.5}   &   \textbf{76.1} & 74.5 & \textbf{86.4} & \textbf{80.6}   \\
\bottomrule
\end{tabular}}
\caption{ Results on multi-task learning using the MixLoRA and OMoE. The backbone model is LLaMA-3 8B.} 
\label{multi-task_llama3}
\vspace{8pt}
\end{table}

\begin{table}[tb!]
    \centering
{
    \begin{tabular}{cccccccc}
        \toprule
 \textbf{Rank $r$}  & \textbf{ARC-e} & \textbf{ARC-c} & \textbf{BoolQ} & \textbf{PIQA} & \textbf{Average}  \\ 
        \midrule
 2  & 73.4  & 50.5  & 72.8  & 82.4  & 69.8   \\
 4 & 75.3  & 51.6  & 73.7  & 83.7  & 71.1 \\ 
 8  & 77.2 & 52.1 & 73.1 & 83.8 & 71.6  \\ 
 16  & 79.3 & 56.6 & 73.5 & 84.5 & \textbf{73.5} \\
32  & 79.7 & 57.7 & 72.5 & 82.8 & 73.2 \\ 
        \bottomrule
    \end{tabular}}
    \caption{Accuracy comparison of OMoE-LoRA with varying ranks for LLaMA-2 7B on various benchmarks.}
    \label{table:robustnessr}
    \vspace{8pt}
\end{table}

\subsection{Main Results}
\textbf{Single-Task Setup.} In this setup, we compare OMoE with the performance of LoRA, DoRA and MixLoRA by employing these methods for fine-tuning a single task. The comparison with other baselines is provided in the Appendix~\ref{Results_other_baselines}. The experimental results on diverse datasets are presented in Table~\ref{table:single_task_results}. We provide the number of tunable parameters in the third column, which shows that OMoE significantly reduces the number of experts and tunable parameters by approximately 75\%. Table~\ref{table:single_task_results} also reveals that OMoE outperforms the baselines in most tasks, with other tasks achieving respectable performance. Additionally, in most datasets, the performance gap between OMoE-LoRA and OMoE-DoRA is smaller than that between LoRA and DoRA. This indicates that OMoE is not sensitive to underlying fine-tuning methods and effectively utilizes the model's capacity.

\textbf{Multi-Task Setup.}
Table~\ref{multi-task} presents the results of LoRA, DoRA, MoELoRA, MiLoRA, MixLoRA and OMoE with LLaMA-2 7B in multi-task learning. In contrast to the single-task setup in Table ~\ref{table:single_task_results}, during multi-task learning, we mix training data from ARC, BoolQ, OBQA, and PIQA to train the model, followed by separate evaluations to investigate the generalization ability of each method. The results indicate that, compared to single-task learning, LoRA and DoRA exhibit degradation in average accuracy in multi-task learning. In this configuration, MoELoRA and MiLoRA exhibit comparable performance, showing slight degradation. MixLoRA benefits from top-2 routing with 8 experts, outperforming the single-task setup, albeit at the cost of requiring more parameters. In contrast, OMoE achieves stable performance improvements under minimal parameter conditions. We also add experiments under a multi-task setting on LLaMA-3 8B, with the relevant results shown in the Table~\ref{multi-task_llama3}.

\subsection{Ablation Studies and Further Analysis}

\begin{figure}[b]  
\centering
\subfigure[Boolq]{
\begin{minipage}[b]{0.22\textwidth}
\includegraphics[width=1\textwidth]{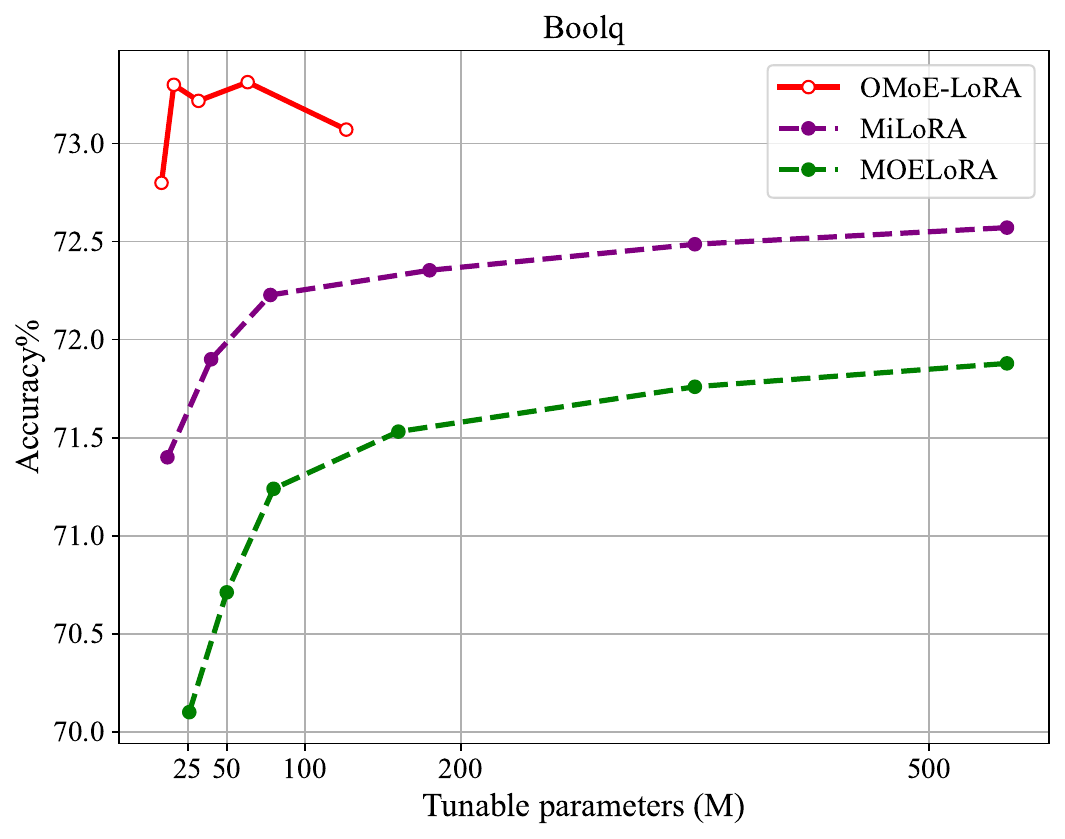} 
\end{minipage}
}
\subfigure[PIQA]{
\begin{minipage}[b]{0.22\textwidth}
\includegraphics[width=1\textwidth]{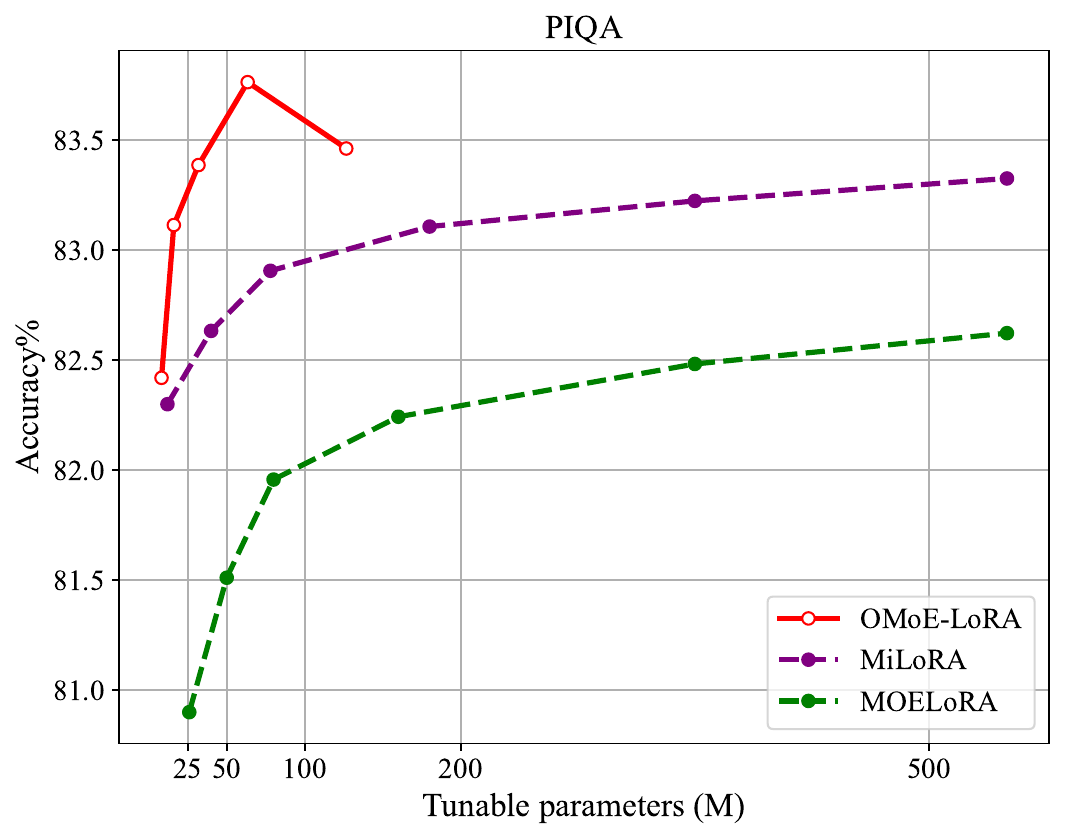} 
\end{minipage}
}
\caption{The performance of OMoE under different numbers of tunable parameters.}
\label{fig:tunable_patameters}
\end{figure}

\textbf{Comparisons under Different Budgets of Tunable Parameters.}
We vary the budget of tunable parameters for OMoE-LoRA by adjusting the values of LoRA rank $r=16$ to $\{2,4,8,32\}$. We also modify the tunable parameters of MiLoRA and MoELoRA as baselines, consistent with ~\cite{zhang2024milora}. The experimental results on the BoolQ and PIQA tasks are presented in Fig.~\ref{fig:tunable_patameters}. Furthermore, we can verify the robustness of OMoE under varying ranks, which is illustrated in Table~\ref{table:robustnessr}. To achieve a better trade-off between performance and training costs, we set the hyperparameter $r$ to 16. The results indicate that under varying budgets of tunable parameters, OMoE-LoRA consistently outperforms the baselines and demonstrates resource efficiency by significantly reducing tunable parameters. As $r$ increases from 2 to 4, 8, and 16, the model's accuracy benefits from the continuous increase in parameters. When $r=32$, the performance of OMoE declines due to the difficulty in convergence under limited training steps.

\begin{figure}[b]  
\centering
\subfigure[Soft Routing]{
\begin{minipage}[b]{0.22\textwidth}
\includegraphics[width=1\textwidth]{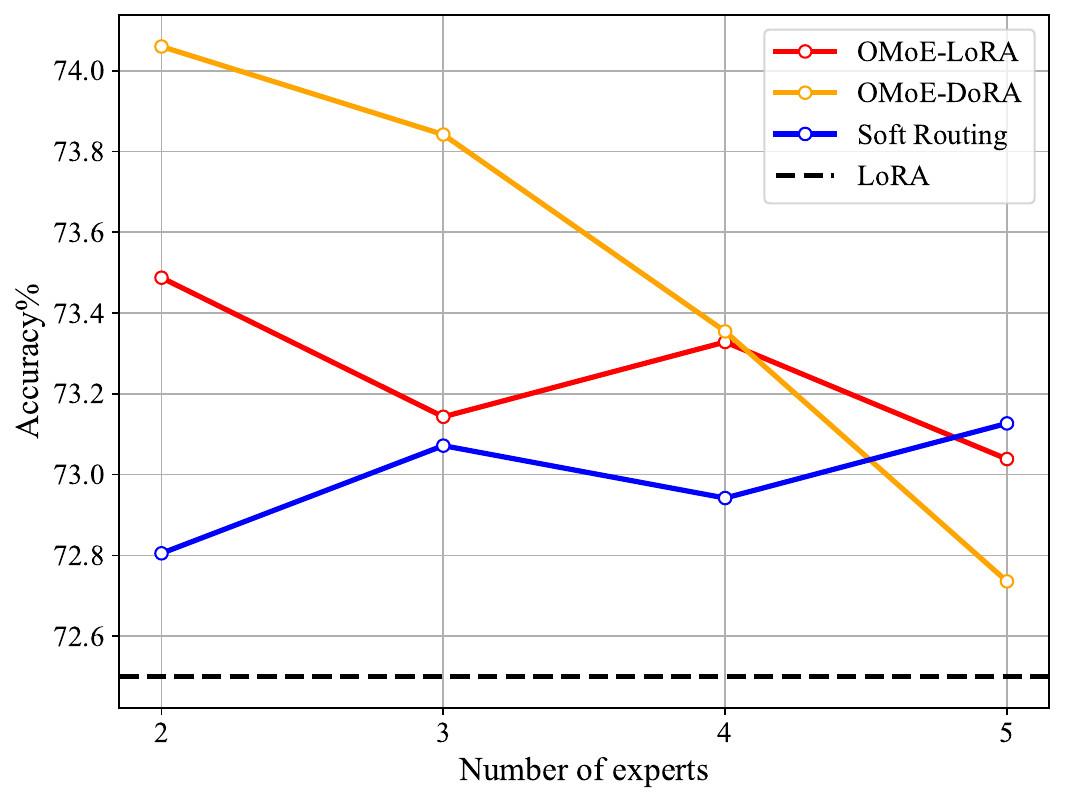} 
\end{minipage}
}
\subfigure[Top-2 Routing]{
\begin{minipage}[b]{0.22\textwidth}
\includegraphics[width=1\textwidth]{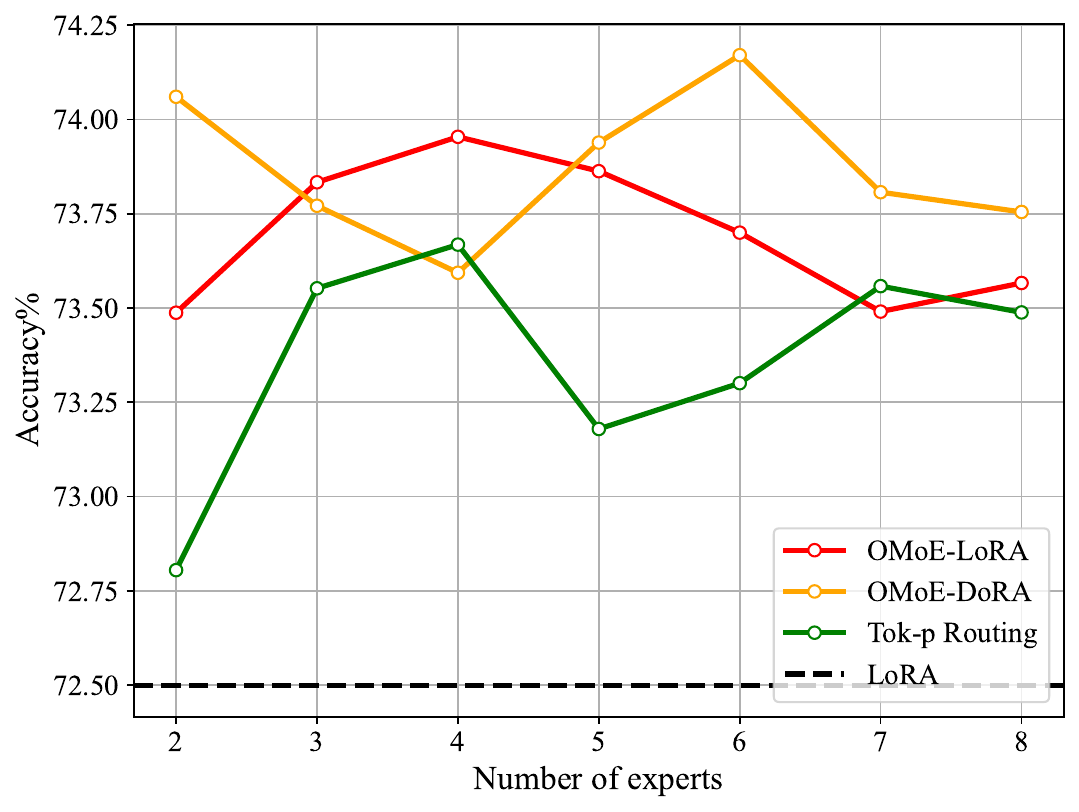} 
\end{minipage}
}
\caption{The performance of OMoE with different numbers of experts using soft routing and top-2 routing.}
\label{fig:differentk}
\end{figure}

\textbf{Effect of k and Routing Strategies.}
In Tables~\ref{table:single_task_results} and ~\ref{tab:results_main_multi_task}, we set the number of LoRA experts, k, to 2. To investigate the effect of $k$, we vary $k$ to \{3, 4, 5, 6, 7\} and explore the performance of OMoE under different routing strategies. For comparison, we use Soft Routing and Top-2 Routing as baselines to validate the performance improvements brought by OMoE. The average performance results for the multi-task setting ([ARC-e, ARC-c, BoolQ, OBQA]) are presented in Fig.~\ref{fig:differentk}. The results indicate that:

(a) Our OMoE method consistently outperforms both LoRA and vanilla MoE routing strategies. 

(b) As shown in Fig.~\ref{fig:differentk}(a), OMoE with Soft Routing exhibits a declining trend as the number of experts increases. The underlying reason is that the weighted averaging mechanism of soft routing proves incompatible with specialized and diversified expert modules. The performance degradation of OMoE in soft routing precisely corroborates OMoE's effectiveness in promoting expert diversity. The robustness of soft routing when varying the number of experts demonstrates the existence of representational redundancy in vanilla MoE.

(c) As illustrated in Fig.~\ref{fig:differentk}(b), our OMoE also benefits slightly from Top-2 routing, demonstrating that expert representation diversity enhances model performance. Furthermore, OMoE substantially reduces tunable parameters while maintaining expert diversity, allowing it to effectively integrate with various routing strategies.

\begin{figure*}[ht!]  
\centering
\subfigure[2 experts]{
\begin{minipage}[b]{0.27\textwidth}
\includegraphics[width=1\textwidth]{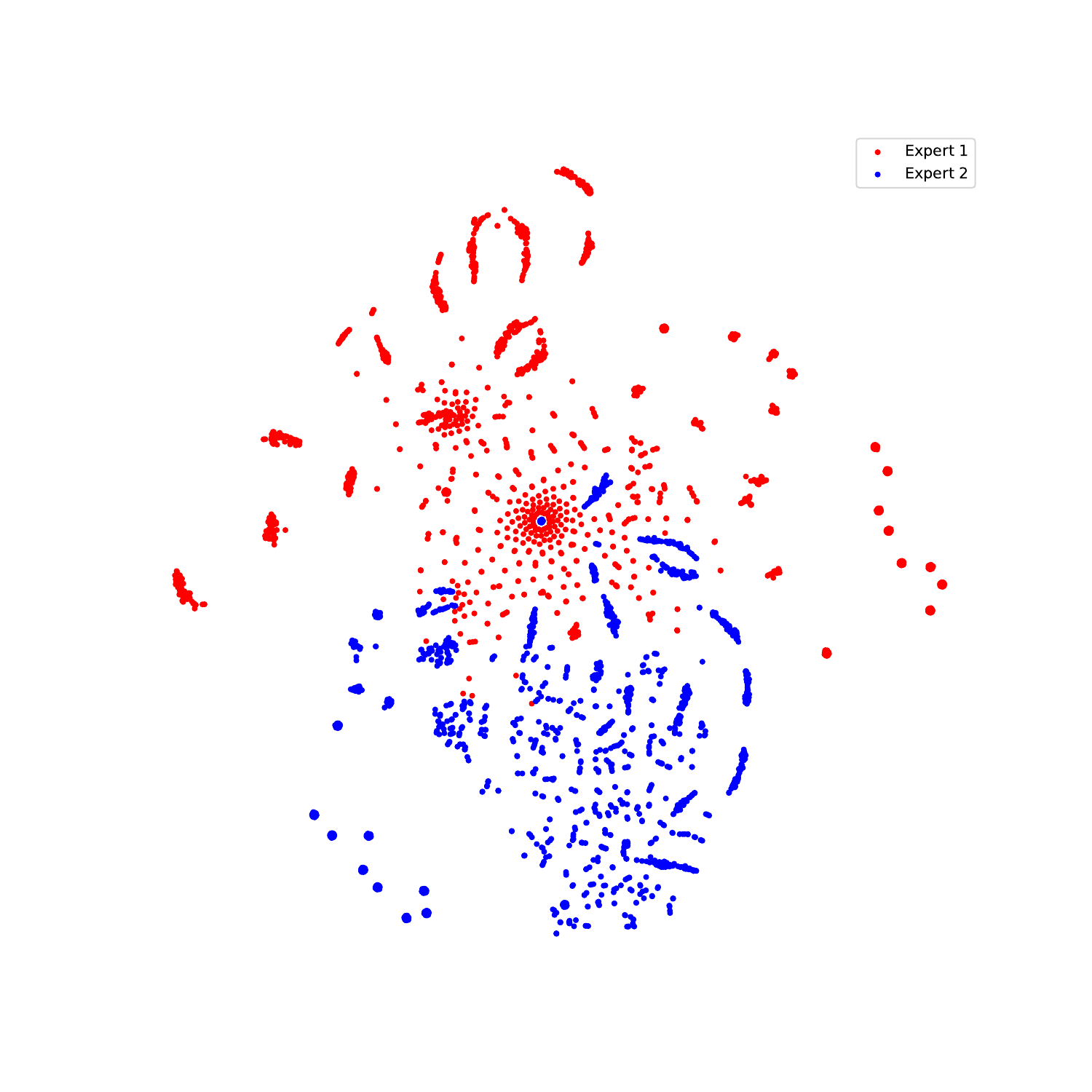} 
\end{minipage}
}
\subfigure[3 experts]{
\begin{minipage}[b]{0.27\textwidth}
\includegraphics[width=1\textwidth]{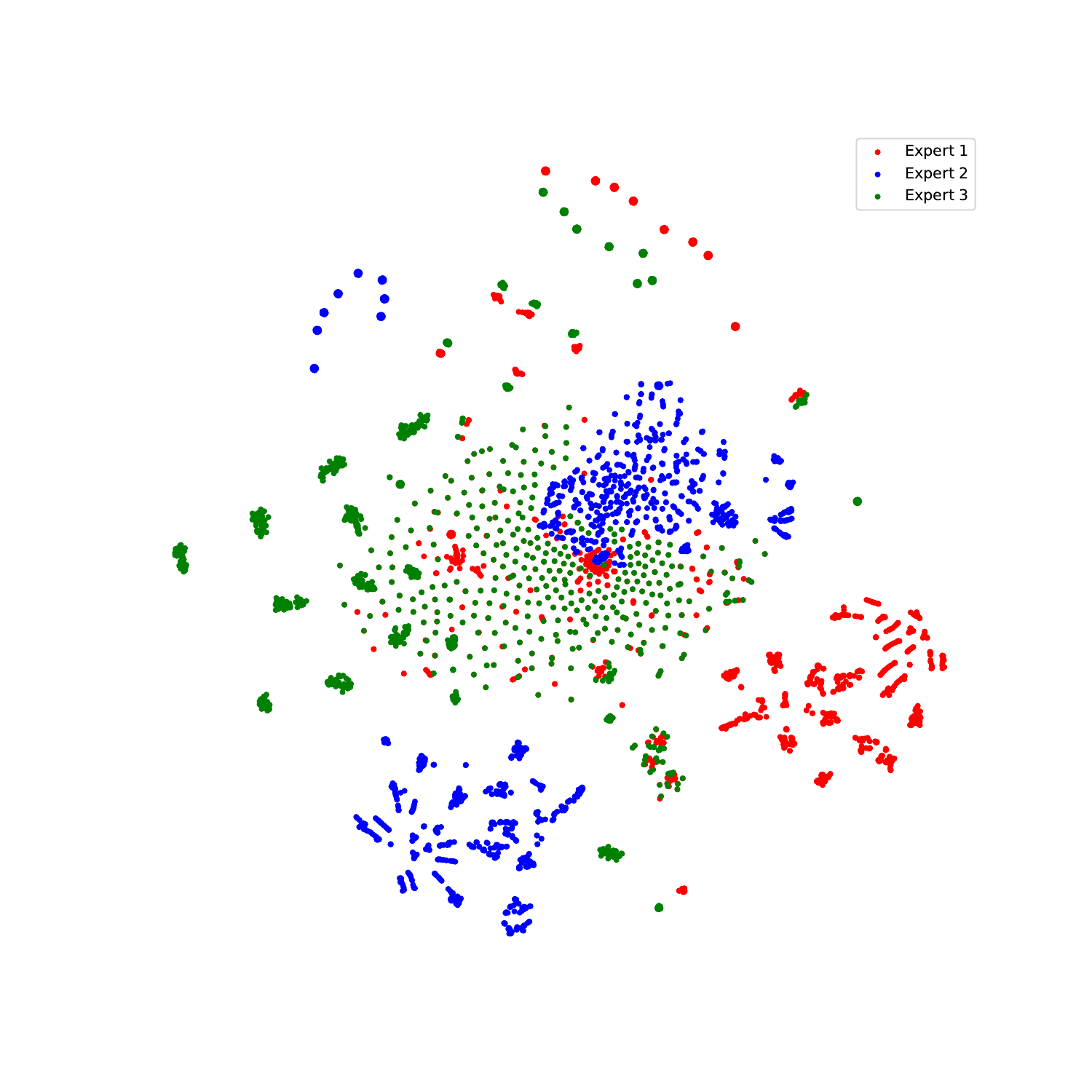} 
\end{minipage}
}
\subfigure[4 experts]{
\begin{minipage}[b]{0.27\textwidth}
\includegraphics[width=1\textwidth]{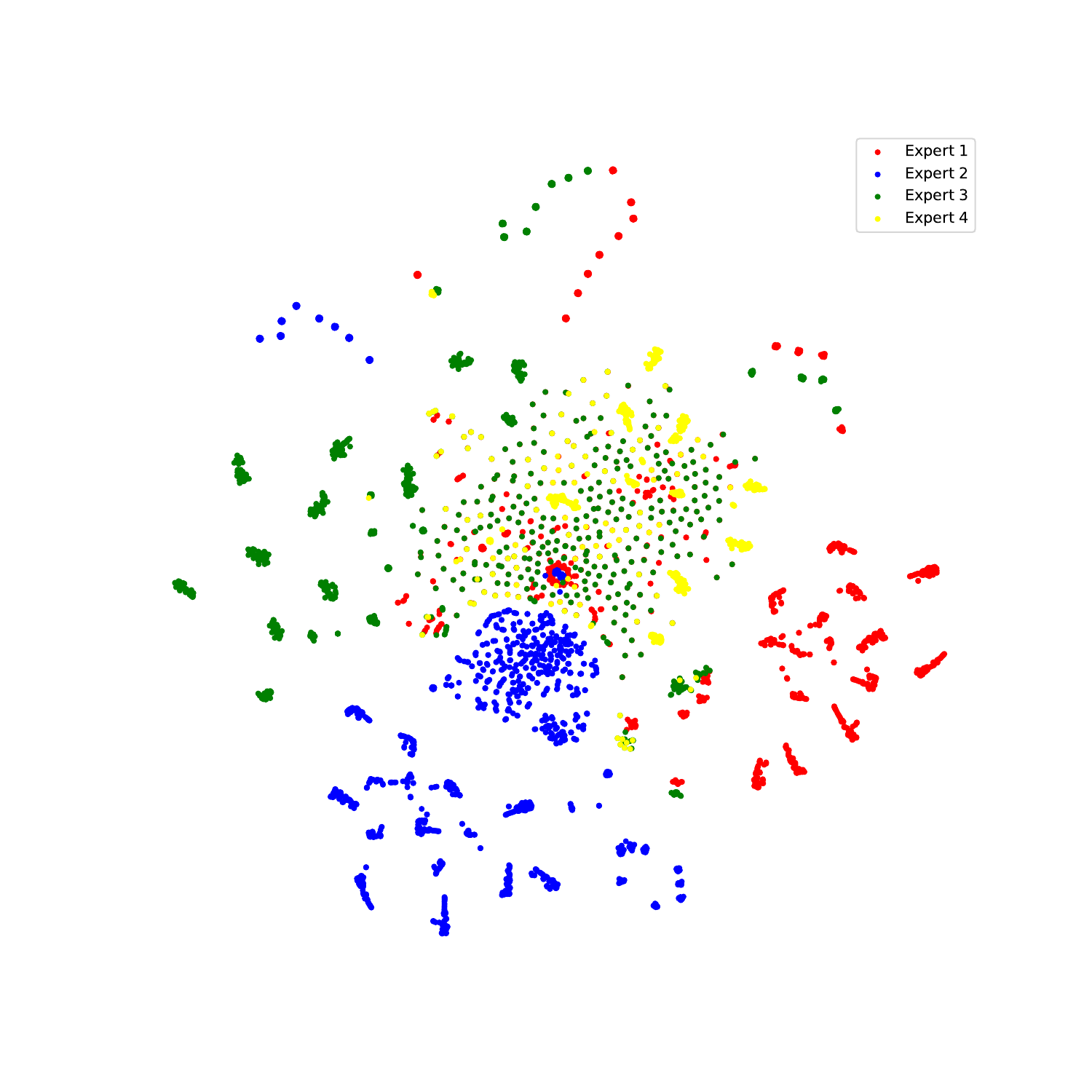} 
\end{minipage}
}
\caption{The representations under different numbers of experts in OMoE using top-2 routing are visualized via t-SNE. The representations are obtained from layer 16 of LLaMA-2 7B. The visualization of more layers is provided in Appendix~\ref{differenk}.}
\vspace{8pt}
\label{fig:visualization_differentk}
\end{figure*}

\textbf{Visulization of Expert Representations.}
OMoE can be regarded as a direct solution for achieving expert diversity in MoE architectures, accomplished by enforcing representation orthogonality. To visualize expert diversity, we conducted fine-tuning experiments on the OBQA dataset: gradually increasing the number of experts in OMoE under top-k routing while visualizing representations from randomly selected expert layers in large models. As illustrated in Fig.~\ref{fig:visualization_differentk}, the visualization results demonstrate OMoE's successful achievement of expert diversity. Furthermore, as shown in Fig.~\ref{fig:differentk}(a), using two orthogonal experts achieves the optimal balance simultaneously achieving expert diversity while maintaining a minimal number of experts. We also visualized representations across different layers in LLMs and please refer to the Appendix~\ref{differentmethod} for more details.

\textbf{Analysis of Model Efficiency.}
To demonstrate the inference efficiency of our OMoE method, we compared the inference latency, GPU memory utilization and training time of LoRA, DoRA, MixLoRA, and OMoE under two typical routing strategies on a batch of input instructions. We assess efficiency based on three metrics: (a) the inference time required for generating responses (ms). (b) the GPU memory cost. (c) the training time for multi-task setting. The results are presented in Table~\ref{efficiency}.

As illustrated in Table~\ref{efficiency}, OMoE (Soft Routing) significantly reduces memory cost and inference latency compared to the well-performing MixLoRA. Specifically, OMoE (Soft Routing) is 30.1\% faster than MixLoRA while decreasing GPU memory cost by 698 MiB. In terms of training time, OMoE requires 2.1 hours on the RTX A6000 in the multi-task setting (arc-e, arc-c, boolq, obqa), while MixLoRA takes 2.2 hours. Although OMoE (Top-2 Routing) slightly increases computational overhead, it slightly enhances models' performance. The advantages of OMoE (Soft Routing) and the analysis can be summarized as follows:
\begin{itemize}
    \item OMoE (Soft Routing) reduces the number of LoRA experts to modify Transformer layers at each decoding step, making generating new tokens more efficient. Compared to OMoE (Top-2 Routing), OMoE merely calls two experts, so even with the additional computational steps for orthogonality, the inference efficiency is still improved.
    \item When comparing OMoE (Top-2 Routing) to MixLoRA, it is clear that although the Gram-Schmidt process incurs additional computational overhead, the reduction in the number of experts during the mapping operation provides significantly greater benefits.
    \item The orthogonalization in OMoE does slightly increase the computational overhead, but compared to MoE methods like MixLoRA, OMoE does not require training a router and does not need to calculate the router balancing loss, which reduces the computational burden. Overall, OMoE reduces the training time.
\end{itemize}

\begin{table}[tb!]
\centering
{
\begin{tabular}{ccccccccc}
\toprule
{\textbf{Method}}   &   \textbf{Latency (ms)}   &   {\textbf{Memory (MiB)} }  &   {\textbf{Training time (h)} }    \\
\hline

{LoRA }  &     2,096   &  +1,630  &1.8h\\    
\hdashline
{DoRA}   &    1,748   &   +2,184  &1.7h \\
\hdashline
{MixLoRA}  &    4,217    &    +1,776 & 2.2h   \\
\hdashline
{OMoE (Soft)  }  &    3,401     &    +1,078 & 2.1h   \\
\hdashline
{OMoE (Top-2)}  &    4,863    &    +1,776 &  2.3h \\
\bottomrule
\end{tabular}}
\caption{The inference latency and memory cost of the LLaMA-2 7B model for generating a batch of responses using different baselines.} 
\vspace{8pt}
\label{efficiency}
\end{table}

\textbf{Analysis of Layer-Wise Orthogonality.}
After demonstrating that orthogonal experts possess a superior capability for handling fine-grained information, a natural question arises: "Which layers in LLMs require orthogonal experts the most?" To conduct a more detailed analysis of the role of orthogonal experts across different layers in LLMs, we take LLaMA-2 7B as an example and simplify the model into three levels: low (from layer 1 to 10), medium (from layer 11 to 20), and high (from layer 21 to 32), injecting OMoE into these levels. Specifically, we propose four types of layer-wise orthogonal expert configurations:
\begin{itemize}
\item \textbf{OMoE-$\bigtriangleup$} injects OMoE into low levels while medium and high levels are vanilla MoE.
\item \textbf{OMoE-$\bigtriangledown$} injects OMoE into high levels while low and medium levels are vanilla MoE.
\item  \textbf{OMoE-$\diamond$} injects OMoE into medium levels while low and high levels are vanilla MoE.
\item  \textbf{OMoE-$\bowtie$} injects OMoE into low and high levels while medium levels is vanilla MoE.
\end{itemize}

\begin{table}[tb!]
\centering
{
\begin{tabular}{cccccc}
\toprule
{\textbf{Method}}   &   \textbf{ARC-e}   &  \textbf{ARC-c}   & \textbf{BoolQ}   & {\textbf{OBQA} } & {\textbf{Avg.} }      \\
\hline

{OMoE-$\bigtriangleup$ }  &     81.3   &  56.4 & 73.4 & 80.2 & 72.8   \\    
\hdashline
{OMoE-$\bigtriangledown$}   &    80.0   &   57.1 & 73.1 & 79.8 & 72.5   \\
\hdashline
{OMoE-$\diamond$}  &    80.1    &    55.1 & 73.9 & 81.0 & 72.5   \\
\hdashline
{OMoE-$\bowtie$}  & 78.6 & 54.9 & 73.1 & 81.0 & 71.9     \\
\bottomrule
\end{tabular}}
\caption{Compare different configurations of layer-wise orthogonal experts for single-task learning.} 
\label{layerwise_single}
\vspace{8pt}
\end{table}

\begin{table}[tb!]
\centering
{
\begin{tabular}{cccccc}
\toprule
{\textbf{Method}}   &   \textbf{ARC-e}   &  \textbf{ARC-c}   & \textbf{BoolQ}   & {\textbf{OBQA} } & {\textbf{Avg.} }      \\
\hline

{OMoE-$\bigtriangleup$ }  &     82.6   &  67.2 & 71.9 & 78.8 & 75.1   \\    
\hdashline
{OMoE-$\bigtriangledown$}   &    80.5   &   67.0 & 72.9 & 79.4 & 74.9   \\
\hdashline
{OMoE-$\diamond$}  &    81.1    &    68.4 & 73.6 & 80.6 & 75.9   \\
\hdashline
{OMoE-$\bowtie$}  &     82.0    &    66.3 & 72.1 & 81.0 & 75.4  \\
\bottomrule
\end{tabular}}
\caption{Compare different configurations of layer-wise orthogonal experts for multi-task learning.} 
\vspace{8pt}
\label{layerwise_multi}
\end{table}
Table~\ref{layerwise_single} shows the results for four types of layer-wise configurations in single-task settings. Our results indicate that in single-task settings, OMoE-$\bowtie$ performs poorly compared to other configurations, while the other three configurations exhibit similar performance. Notably, OMoE-$\bigtriangleup$ achieves the best results among them, suggesting that injecting OMoE into the lowest layers is more advantageous for capturing task-relevant information, thereby enhancing the model's fine-tuning effectiveness.

Table~\ref{layerwise_multi} presents the results for four types of layer-wise configurations in multi-task settings. Our results indicate that OMoE-$\diamond$ achieves a slight performance improvement compared to OMoE-$\bowtie$ while requiring less computational resources. Conversely, OMoE-$\bigtriangledown$ and OMoE-$\bigtriangleup$ are unable to attain comparable performance due to the lack of OMoE in the medium levels. From the experiments, we can conclude that injecting OMoE into the medium layers is more beneficial for handling heterogeneous data.

\subsection{Discussion: Redundancy in Visualization }
As illustrated in Fig.~\ref{fig:visualization_differentk}, when $k=2$, the experts within OMoE benefit from orthogonal constraints, leading to diversified representations. However, the increase in the number of experts ($k \geq 3$) inevitably introduces redundancy. To explain this phenomenon, we propose the following interpretation: The similar representations are the result of the final optimization objective, that is, in the learning process, the objective function being optimized forces the latter orthogonal experts to output representations similar to those of the previous orthogonal experts. For example: Firstly, we have two learned expert representations $(v_1=(1,0,0)$ and $(v_2=(0,0,5)$. Then, the newly added orthogonal expert $(v_3=(0,0.9,0)$ attempts to optimize along dim=2, but the final optimization objective forces it to approach the meaningful representation vector $(v_1=(1,0,0)$. Therefore, although the experts are orthogonal, expert 3 is still similar to expert 1 in the visualization, resulting in redundancy. This redundancy is caused by the optimization objective, which suppresses the diversity of more experts, rather than the inherent redundancy of the MoE.

\section{Related Works}
\label{relatedworks}
\textbf{Parameter-Efficient Fine-Tuning}: To enhance the performance of LLMs for downstream tasks, fine-tuning is an unavoidable step. However, direct full-parameter fine-tuning is  time-consuming and memory-intensive, posing challenges for practical applications. Consequently, researchers are motivated to explore parameter-efficient fine-tuning (PEFT) techniques aimed at reducing the time and memory costs during fine-tuning. The core idea of Parameter-Efficient Fine-Tuning (PEFT) is to fine-tune only a small subset of model parameters while keeping the majority of pre-trained parameters unchanged. Representative methods include Adapters, Prefix-tuning, and Prompt-tuning. Adapters~\cite{zhang2023learned,he2021towards,hu2022sparse}, one of the earliest approaches, improve the adaptability of large models by optimizing their architecture through automatic search. BitFit~\cite{zaken2021bitfit} and $(\text{IA})^3$~\cite{liu2022few} link learnable vectors to the hidden states of various modules within the Transformer layers using either multiplication or addition, thereby facilitating efficient fine-tuning. Prefix-tuning~\cite{li2021prefix} adds embeddings to the hidden states of the Transformer model, while Prompt-tuning~\cite{liu2021p,zhu2024iapt} only adds embeddings to the input layer. LoRA~\cite{hu2021lora}, which utilizes low-rank matrix decomposition of linear layer weights, has become one of the popular PEFT methods, with many works dedicated to improving LoRA. AdaLoRA~\cite{zhang2023adalora} investigates the parameter allocation of LoRA modules. DoRA~\cite{liu2024dora} decomposes pre-trained weights into magnitude and direction components to reduce the number of trainable parameters. Lora+~\cite{hayou2024loraplus} sets different but fixed learning rates for the adapter matrices A and B in LoRA to correct the suboptimalities of LoRA. rsLoRA~\cite{kalajdzievski2023rank} modifies LoRA by using appropriate scaling factors to achieve a computation/performance trade-off during fine-tuning. Vera~\cite{kopiczko2023vera}  randomly freezes initialized LoRA matrices and only fine-tunes a set of scaling vectors.

\textbf{MoE Variants}: MoE variants utilize a combination of parallel LoRA modules, referred to as "experts", and a gating mechanism that selects which experts to activate based on the input data. This architecture achieves a better trade-off between performance and training cost, making it a popular choice for efficiently fine-tuning large language models. Mixlora~\cite{li2024mixlora} and Moelora~\cite{liu2023moelora} integrate Low-Rank Adaptation (LoRA) with MoE for multi-task learning, constructing a resource-efficient sparse model. Loramoe~\cite{dou2023loramoe} incorporates a localized balancing constraint, which enables two expert groups to maintain specialized functions between groups while achieving balanced workloads within each group. Llava-mole~\cite{chen2024llava} adopts a sparse combination of LoRA experts to address data conflicts when handling instruction data from different domains. SCMoE~\cite{shi2024unchosen} leverages contrastive learning within an MoE framework to enhance parameter efficiency during fine-tuning processes for LLMs. HydraLoRA~\cite{tian2024hydralora} introduces an asymmetric structure to the LoRA framework without requiring domain expertise, improving both training and inference performance and efficiency. MiLoRA~\cite{zhang2024milora} employs a prompt-aware routing mechanism, allowing it to reuse these results for subsequent tokens, thereby reducing latency. Teamlora~\cite{lin2024teamlora} emphasizes collaboration and competition among experts, further enhancing the effectiveness of low-rank adaptation. Some works apply MoE architectures to the inference stage of large models to achieve better generalization. LoraHub~\cite{huang2023lorahub} and LLMAdapters~\cite{hu2023llm} adopt the MoE structure by training multiple adapters and strategically selecting adapter combinations based on the domain during inference, thereby improving the model's performance in multi-task environments.

\section{Conclusion and Discussion}
This paper investigates the expert collapse in MoE. We analyze the benefits of diversity in MoE and develop OMoE, a resource-efficient MoE variant designed to promote expert diversity, which facilitates stable and efficient performance improvements. In OMoE, the Gram-Schmidt process is leveraged to ensure that the experts' representations lie within the Stiefel manifold. This process enables OMoE to reduce representation redundancy when fine-tuning MoE. Additionally, OMoE keeps the learning objective unchanged by imposing constraints. Our qualitative analysis and experimental results highlight the benefits of OMoE in efficient fine-tuning for LLMs. Based on the performance of OMoE on supervised fine-tuning, we believe the higher potential of OMoE can be exploited with Reinforcement Learning from Human Feedback (RLHF) in future study.



\begin{ack}
This work was supported in part by the National Natural Science Foundation of China under Grant 62322316, the Key Research Project of Chinese Academy of Sciences under Grant No. KGFZD-145-25-21-01, the Open Fund of National Key Laboratory of Information Systems Engineering (No. 6142101240203), and the Beijing Nova Program under Grant 20220484077 and 20230484435.
\end{ack}
\clearpage


\bibliography{main}

\begin{thebibliography}{47}
\providecommand{\natexlab}[1]{#1}
\providecommand{\url}[1]{\texttt{#1}}
\expandafter\ifx\csname urlstyle\endcsname\relax
  \providecommand{\doi}[1]{doi: #1}\else
  \providecommand{\doi}{doi: \begingroup \urlstyle{rm}\Url}\fi

\bibitem[Achiam et~al.(2023)Achiam, Adler, Agarwal, Ahmad, Akkaya, Aleman, Almeida, Altenschmidt, Altman, Anadkat, et~al.]{achiam2023gpt}
J.~Achiam, S.~Adler, S.~Agarwal, L.~Ahmad, I.~Akkaya, F.~L. Aleman, D.~Almeida, J.~Altenschmidt, S.~Altman, S.~Anadkat, et~al.
\newblock Gpt-4 technical report.
\newblock \emph{arXiv preprint arXiv:2303.08774}, 2023.

\bibitem[Bisk et~al.(2020)Bisk, Zellers, Gao, Choi, et~al.]{bisk2020piqa}
Y.~Bisk, R.~Zellers, J.~Gao, Y.~Choi, et~al.
\newblock Piqa: Reasoning about physical commonsense in natural language.
\newblock In \emph{Proceedings of the AAAI conference on artificial intelligence}, volume~34, pages 7432--7439, 2020.

\bibitem[Chan(2023)]{chan2023gpt}
A.~Chan.
\newblock Gpt-3 and instructgpt: technological dystopianism, utopianism, and “contextual” perspectives in ai ethics and industry.
\newblock \emph{AI and Ethics}, 3\penalty0 (1):\penalty0 53--64, 2023.

\bibitem[Chen et~al.(2024)Chen, Jie, and Ma]{chen2024llava}
S.~Chen, Z.~Jie, and L.~Ma.
\newblock Llava-mole: Sparse mixture of lora experts for mitigating data conflicts in instruction finetuning mllms.
\newblock \emph{arXiv preprint arXiv:2401.16160}, 2024.

\bibitem[Chi et~al.(2022)Chi, Dong, Huang, Dai, Ma, Patra, Singhal, Bajaj, Song, Mao, et~al.]{chi2022representation}
Z.~Chi, L.~Dong, S.~Huang, D.~Dai, S.~Ma, B.~Patra, S.~Singhal, P.~Bajaj, X.~Song, X.-L. Mao, et~al.
\newblock On the representation collapse of sparse mixture of experts.
\newblock \emph{Advances in Neural Information Processing Systems}, 35:\penalty0 34600--34613, 2022.

\bibitem[Clark et~al.(2019)Clark, Lee, Chang, Kwiatkowski, Collins, and Toutanova]{clark2019boolq}
C.~Clark, K.~Lee, M.-W. Chang, T.~Kwiatkowski, M.~Collins, and K.~Toutanova.
\newblock Boolq: Exploring the surprising difficulty of natural yes/no questions.
\newblock \emph{arXiv preprint arXiv:1905.10044}, 2019.

\bibitem[Clark et~al.(2018)Clark, Cowhey, Etzioni, Khot, Sabharwal, Schoenick, and Tafjord]{clark2018think}
P.~Clark, I.~Cowhey, O.~Etzioni, T.~Khot, A.~Sabharwal, C.~Schoenick, and O.~Tafjord.
\newblock Think you have solved question answering? try arc, the ai2 reasoning challenge.
\newblock \emph{arXiv preprint arXiv:1803.05457}, 2018.

\bibitem[Dou et~al.(2023)Dou, Zhou, Liu, Gao, Zhao, Shen, Zhou, Xi, Wang, Fan, et~al.]{dou2023loramoe}
S.~Dou, E.~Zhou, Y.~Liu, S.~Gao, J.~Zhao, W.~Shen, Y.~Zhou, Z.~Xi, X.~Wang, X.~Fan, et~al.
\newblock Loramoe: Revolutionizing mixture of experts for maintaining world knowledge in language model alignment.
\newblock \emph{arXiv preprint arXiv:2312.09979}, 4\penalty0 (7), 2023.

\bibitem[Hayou et~al.(2024)Hayou, Ghosh, and Yu]{hayou2024loraplus}
S.~Hayou, N.~Ghosh, and B.~Yu.
\newblock Lora+: Efficient low rank adaptation of large models.
\newblock \emph{arXiv preprint arXiv:2402.12354}, 2024.

\bibitem[He et~al.(2021)He, Zhou, Ma, Berg-Kirkpatrick, and Neubig]{he2021towards}
J.~He, C.~Zhou, X.~Ma, T.~Berg-Kirkpatrick, and G.~Neubig.
\newblock Towards a unified view of parameter-efficient transfer learning.
\newblock \emph{arXiv preprint arXiv:2110.04366}, 2021.

\bibitem[He et~al.(2015)He, Zhang, Ren, and Sun]{he2015delving}
K.~He, X.~Zhang, S.~Ren, and J.~Sun.
\newblock Delving deep into rectifiers: Surpassing human-level performance on imagenet classification.
\newblock In \emph{Proceedings of the IEEE international conference on computer vision}, pages 1026--1034, 2015.

\bibitem[Hendawy et~al.(2023)Hendawy, Peters, and D'Eramo]{hendawy2023multi}
A.~Hendawy, J.~Peters, and C.~D'Eramo.
\newblock Multi-task reinforcement learning with mixture of orthogonal experts.
\newblock \emph{arXiv preprint arXiv:2311.11385}, 2023.

\bibitem[Hu et~al.(2021)Hu, Shen, Wallis, Allen-Zhu, Li, Wang, Wang, and Chen]{hu2021lora}
E.~J. Hu, Y.~Shen, P.~Wallis, Z.~Allen-Zhu, Y.~Li, S.~Wang, L.~Wang, and W.~Chen.
\newblock Lora: Low-rank adaptation of large language models.
\newblock \emph{arXiv preprint arXiv:2106.09685}, 2021.

\bibitem[Hu et~al.(2022)Hu, Zhang, Ding, Wang, Wang, Liu, and Sun]{hu2022sparse}
S.~Hu, Z.~Zhang, N.~Ding, Y.~Wang, Y.~Wang, Z.~Liu, and M.~Sun.
\newblock Sparse structure search for parameter-efficient tuning.
\newblock \emph{arXiv preprint arXiv:2206.07382}, 2022.

\bibitem[Hu et~al.(2023)Hu, Wang, Lan, Xu, Lim, Bing, Xu, Poria, and Lee]{hu2023llm}
Z.~Hu, L.~Wang, Y.~Lan, W.~Xu, E.-P. Lim, L.~Bing, X.~Xu, S.~Poria, and R.~Lee.
\newblock Llm-adapters: An adapter family for parameter-efficient fine-tuning of large language models.
\newblock In \emph{Proceedings of the 2023 Conference on Empirical Methods in Natural Language Processing}, pages 5254--5276, 2023.

\bibitem[Huang et~al.(2023)Huang, Liu, Lin, Pang, Du, and Lin]{huang2023lorahub}
C.~Huang, Q.~Liu, B.~Y. Lin, T.~Pang, C.~Du, and M.~Lin.
\newblock Lorahub: Efficient cross-task generalization via dynamic lora composition.
\newblock \emph{arXiv preprint arXiv:2307.13269}, 2023.

\bibitem[Huang et~al.(2024)Huang, Bai, Zhu, Zhang, Zhang, Su, Liu, Lv, Zhang, Fu, et~al.]{huang2024c}
Y.~Huang, Y.~Bai, Z.~Zhu, J.~Zhang, J.~Zhang, T.~Su, J.~Liu, C.~Lv, Y.~Zhang, Y.~Fu, et~al.
\newblock C-eval: A multi-level multi-discipline chinese evaluation suite for foundation models.
\newblock \emph{Advances in Neural Information Processing Systems}, 36, 2024.

\bibitem[Kalajdzievski(2023)]{kalajdzievski2023rank}
D.~Kalajdzievski.
\newblock A rank stabilization scaling factor for fine-tuning with lora.
\newblock \emph{arXiv preprint arXiv:2312.03732}, 2023.

\bibitem[Kim et~al.(2024)Kim, Shin, Park, and Sung]{kim2024sample}
W.~Kim, Y.~Shin, J.~Park, and Y.~Sung.
\newblock Sample-efficient and safe deep reinforcement learning via reset deep ensemble agents.
\newblock \emph{Advances in Neural Information Processing Systems}, 36, 2024.

\bibitem[Kopiczko et~al.(2023)Kopiczko, Blankevoort, and Asano]{kopiczko2023vera}
D.~J. Kopiczko, T.~Blankevoort, and Y.~M. Asano.
\newblock Vera: Vector-based random matrix adaptation.
\newblock \emph{arXiv preprint arXiv:2310.11454}, 2023.

\bibitem[Lan et~al.(2024)Lan, Zhang, Yi, Guo, Peng, Gao, Wu, Chen, Du, Hu, et~al.]{lan2024contrastive}
S.~Lan, R.~Zhang, Q.~Yi, J.~Guo, S.~Peng, Y.~Gao, F.~Wu, R.~Chen, Z.~Du, X.~Hu, et~al.
\newblock Contrastive modules with temporal attention for multi-task reinforcement learning.
\newblock \emph{Advances in Neural Information Processing Systems}, 36, 2024.

\bibitem[Li et~al.(2024)Li, Ma, Wang, Cheng, Duan, Zuo, Yang, and Tang]{li2024mixlora}
D.~Li, Y.~Ma, N.~Wang, Z.~Cheng, L.~Duan, J.~Zuo, C.~Yang, and M.~Tang.
\newblock Mixlora: Enhancing large language models fine-tuning with lora based mixture of experts.
\newblock \emph{arXiv preprint arXiv:2404.15159}, 2024.

\bibitem[Li et~al.(2020)Li, Fuxin, and Todorovic]{li2020efficient}
J.~Li, L.~Fuxin, and S.~Todorovic.
\newblock Efficient riemannian optimization on the stiefel manifold via the cayley transform.
\newblock \emph{arXiv preprint arXiv:2002.01113}, 2020.

\bibitem[Li and Liang(2021)]{li2021prefix}
X.~L. Li and P.~Liang.
\newblock Prefix-tuning: Optimizing continuous prompts for generation.
\newblock \emph{arXiv preprint arXiv:2101.00190}, 2021.

\bibitem[Lin et~al.(2024)Lin, Liu, Zhang, Li, Dai, Li, Yu, He, Li, Jiang, et~al.]{lin2024teamlora}
T.~Lin, J.~Liu, W.~Zhang, Z.~Li, Y.~Dai, H.~Li, Z.~Yu, W.~He, J.~Li, H.~Jiang, et~al.
\newblock Teamlora: Boosting low-rank adaptation with expert collaboration and competition.
\newblock \emph{arXiv preprint arXiv:2408.09856}, 2024.

\bibitem[Liu et~al.(2023{\natexlab{a}})Liu, Ding, Shen, Peng, Cao, Cheng, and Tao]{liu2023diversifying}
B.~Liu, L.~Ding, L.~Shen, K.~Peng, Y.~Cao, D.~Cheng, and D.~Tao.
\newblock Diversifying the mixture-of-experts representation for language models with orthogonal optimizer.
\newblock \emph{arXiv preprint arXiv:2310.09762}, 2023{\natexlab{a}}.

\bibitem[Liu et~al.(2022)Liu, Tam, Muqeeth, Mohta, Huang, Bansal, and Raffel]{liu2022few}
H.~Liu, D.~Tam, M.~Muqeeth, J.~Mohta, T.~Huang, M.~Bansal, and C.~A. Raffel.
\newblock Few-shot parameter-efficient fine-tuning is better and cheaper than in-context learning.
\newblock \emph{Advances in Neural Information Processing Systems}, 35:\penalty0 1950--1965, 2022.

\bibitem[Liu et~al.(2023{\natexlab{b}})Liu, Wu, Zhao, Zhu, Xu, Tian, and Zheng]{liu2023moelora}
Q.~Liu, X.~Wu, X.~Zhao, Y.~Zhu, D.~Xu, F.~Tian, and Y.~Zheng.
\newblock Moelora: An moe-based parameter efficient fine-tuning method for multi-task medical applications.
\newblock \emph{arXiv preprint arXiv:2310.18339}, 2023{\natexlab{b}}.

\bibitem[Liu et~al.(2024)Liu, Wang, Yin, Molchanov, Wang, Cheng, and Chen]{liu2024dora}
S.-Y. Liu, C.-Y. Wang, H.~Yin, P.~Molchanov, Y.-C.~F. Wang, K.-T. Cheng, and M.-H. Chen.
\newblock Dora: Weight-decomposed low-rank adaptation.
\newblock \emph{arXiv preprint arXiv:2402.09353}, 2024.

\bibitem[Liu et~al.(2021)Liu, Ji, Fu, Tam, Du, Yang, and Tang]{liu2021p}
X.~Liu, K.~Ji, Y.~Fu, W.~L. Tam, Z.~Du, Z.~Yang, and J.~Tang.
\newblock P-tuning v2: Prompt tuning can be comparable to fine-tuning universally across scales and tasks.
\newblock \emph{arXiv preprint arXiv:2110.07602}, 2021.

\bibitem[Mihaylov et~al.(2018)Mihaylov, Clark, Khot, and Sabharwal]{mihaylov2018can}
T.~Mihaylov, P.~Clark, T.~Khot, and A.~Sabharwal.
\newblock Can a suit of armor conduct electricity? a new dataset for open book question answering.
\newblock \emph{arXiv preprint arXiv:1809.02789}, 2018.

\bibitem[Ouyang et~al.(2022)Ouyang, Wu, Jiang, Almeida, Wainwright, Mishkin, Zhang, Agarwal, Slama, Ray, et~al.]{ouyang2022training}
L.~Ouyang, J.~Wu, X.~Jiang, D.~Almeida, C.~Wainwright, P.~Mishkin, C.~Zhang, S.~Agarwal, K.~Slama, A.~Ray, et~al.
\newblock Training language models to follow instructions with human feedback.
\newblock \emph{Advances in neural information processing systems}, 35:\penalty0 27730--27744, 2022.

\bibitem[Qian et~al.(2024)Qian, Xie, Wang, Liu, Dang, Du, Chen, Yang, Liu, and Sun]{qian2024scaling}
C.~Qian, Z.~Xie, Y.~Wang, W.~Liu, Y.~Dang, Z.~Du, W.~Chen, C.~Yang, Z.~Liu, and M.~Sun.
\newblock Scaling large-language-model-based multi-agent collaboration.
\newblock \emph{arXiv preprint arXiv:2406.07155}, 2024.

\bibitem[Qiu et~al.(2023)Qiu, Liu, Feng, Xue, Feng, Liu, Zhang, Weller, and Sch{\"o}lkopf]{qiu2023controlling}
Z.~Qiu, W.~Liu, H.~Feng, Y.~Xue, Y.~Feng, Z.~Liu, D.~Zhang, A.~Weller, and B.~Sch{\"o}lkopf.
\newblock Controlling text-to-image diffusion by orthogonal finetuning.
\newblock \emph{Advances in Neural Information Processing Systems}, 36:\penalty0 79320--79362, 2023.

\bibitem[Sakaguchi et~al.(2021)Sakaguchi, Bras, Bhagavatula, and Choi]{sakaguchi2021winogrande}
K.~Sakaguchi, R.~L. Bras, C.~Bhagavatula, and Y.~Choi.
\newblock Winogrande: An adversarial winograd schema challenge at scale.
\newblock \emph{Communications of the ACM}, 64\penalty0 (9):\penalty0 99--106, 2021.

\bibitem[Sap et~al.(2019)Sap, Rashkin, Chen, LeBras, and Choi]{sap2019socialiqa}
M.~Sap, H.~Rashkin, D.~Chen, R.~LeBras, and Y.~Choi.
\newblock Socialiqa: Commonsense reasoning about social interactions.
\newblock \emph{arXiv preprint arXiv:1904.09728}, 2019.

\bibitem[Shi et~al.(2024)Shi, Yang, Zhu, Wang, Wu, Li, Cai, Yang, and Meng]{shi2024unchosen}
C.~Shi, C.~Yang, X.~Zhu, J.~Wang, T.~Wu, S.~Li, D.~Cai, Y.~Yang, and Y.~Meng.
\newblock Unchosen experts can contribute too: Unleashing moe models' power by self-contrast.
\newblock \emph{arXiv preprint arXiv:2405.14507}, 2024.

\bibitem[Song et~al.(2023)Song, Liu, Chen, An, Zhang, Wang, and Xu]{song2023label}
R.~Song, Z.~Liu, X.~Chen, H.~An, Z.~Zhang, X.~Wang, and H.~Xu.
\newblock Label prompt for multi-label text classification.
\newblock \emph{Applied Intelligence}, 53\penalty0 (8):\penalty0 8761--8775, 2023.

\bibitem[Tian et~al.(2024)Tian, Shi, Guo, Li, and Xu]{tian2024hydralora}
C.~Tian, Z.~Shi, Z.~Guo, L.~Li, and C.~Xu.
\newblock Hydralora: An asymmetric lora architecture for efficient fine-tuning.
\newblock \emph{arXiv preprint arXiv:2404.19245}, 2024.

\bibitem[Wang et~al.(2024)Wang, Zheng, Dai, Yue, Zhu, and Wang]{wang2024ts}
P.~Wang, H.~Zheng, S.~Dai, W.~Yue, W.~Zhu, and X.~Wang.
\newblock Ts-tcd: Triplet-level cross-modal distillation for time-series forecasting using large language models.
\newblock \emph{arXiv preprint arXiv:2409.14978}, 2024.

\bibitem[Zaken et~al.(2021)Zaken, Ravfogel, and Goldberg]{zaken2021bitfit}
E.~B. Zaken, S.~Ravfogel, and Y.~Goldberg.
\newblock Bitfit: Simple parameter-efficient fine-tuning for transformer-based masked language-models.
\newblock \emph{arXiv preprint arXiv:2106.10199}, 2021.

\bibitem[Zellers et~al.(2019)Zellers, Holtzman, Bisk, Farhadi, and Choi]{zellers2019hellaswag}
R.~Zellers, A.~Holtzman, Y.~Bisk, A.~Farhadi, and Y.~Choi.
\newblock Hellaswag: Can a machine really finish your sentence?
\newblock \emph{arXiv preprint arXiv:1905.07830}, 2019.

\bibitem[Zhang et~al.(2024)Zhang, Zhao, Chen, Tian, Zheng, and Zhu]{zhang2024milora}
J.~Zhang, Y.~Zhao, D.~Chen, X.~Tian, H.~Zheng, and W.~Zhu.
\newblock Milora: Efficient mixture of low-rank adaptation for large language models fine-tuning.
\newblock In \emph{Findings of the Association for Computational Linguistics: EMNLP 2024}, pages 17071--17084, 2024.

\bibitem[Zhang et~al.(2023{\natexlab{a}})Zhang, Chen, Bukharin, Karampatziakis, He, Cheng, Chen, and Zhao]{zhang2023adalora}
Q.~Zhang, M.~Chen, A.~Bukharin, N.~Karampatziakis, P.~He, Y.~Cheng, W.~Chen, and T.~Zhao.
\newblock Adalora: Adaptive budget allocation for parameter-efficient fine-tuning.
\newblock \emph{arXiv preprint arXiv:2303.10512}, 2023{\natexlab{a}}.

\bibitem[Zhang et~al.(2023{\natexlab{b}})Zhang, Wang, Tan, and Zhu]{zhang2023learned}
Y.~Zhang, P.~Wang, M.~Tan, and W.~Zhu.
\newblock Learned adapters are better than manually designed adapters.
\newblock In \emph{Findings of the Association for Computational Linguistics: ACL 2023}, pages 7420--7437, 2023{\natexlab{b}}.

\bibitem[Zhu et~al.(2024)Zhu, Tian, Yin, Ni, Wang, and Xie]{zhu2024iapt}
W.~Zhu, A.~X. Tian, C.~Yin, Y.~Ni, X.~Wang, and G.~Xie.
\newblock Iapt: Instruction-aware prompt tuning for large language models.
\newblock \emph{arXiv preprint arXiv:2405.18203}, 2024.

\bibitem[Zhuang et~al.(2023)Zhuang, Yu, Wang, Sun, and Zhang]{zhuang2023toolqa}
Y.~Zhuang, Y.~Yu, K.~Wang, H.~Sun, and C.~Zhang.
\newblock Toolqa: A dataset for llm question answering with external tools.
\newblock \emph{Advances in Neural Information Processing Systems}, 36:\penalty0 50117--50143, 2023.

\end{thebibliography}
\clearpage
\appendix
\begin{center}
\textbf{\LARGE Appendix}
\end{center}

\section{Datasets}
\label{sec:appendix_datasets}
The detailed information about the datasets used in the experiments is presented in Table \ref{tab:dataset_stats}. All datasets are downloaded from Hugging Face.

\begin{table*}[ht!]
\centering
{
\begin{tabular}{cccccc}
\hline
Datasets  &  \#train      &   \#test   &     Type   &   Metrics  \\ 
\hline
BoolQ  &     9,427	   &   3,270   &  Text Classification   &  acc    \\
OBQA    &  4,957	      &   500     &  Question Answering   &  acc      \\
ARC-e   &    2,251	  &   2,376    &  Question Answering   &  acc      \\
ARC-c   &    1,119	  &   1,172    &  Question Answering   &  acc      \\
PIQA   &   16,100      &    1,840   &  Question Answering   &  acc \\
SIQA   &   33,410      &    1,954   &  Question Answering   &  acc \\
HellaSwag   &   39,905      &    10,042   &  Sentence Completion   &  acc \\
WinoGrande   &   9,248      &    1,267   &  Fill in the Blank   &  acc \\

\hline
\end{tabular}}
\caption{\label{tab:dataset_stats}  The dataset statistics. }
\end{table*}

\section{Experimental settings}
\label{sec:experimental settings}
\begin{table}[h]
\caption{Hyperparameter configurations of OMoE for fine-tuning LLaMA2-7B/13B, and LLaMA3-8B on datasets. }
\label{table:hyperparameters}
\centering
\begin{tabular}{lc}
\toprule
\textbf{Hyperparameters}       & \textbf{OMoE-LoRA/DoRA}   \\
\midrule
Cutoff Length &{512}\\
Learning Rate &{2e-4}\\
Optimizer     &{AdamW}\\
Batch size    &{16}\\
Accumulation Steps & {8}\\
Dropout       &{0.05}\\
\# Epochs        &{2}\\
Where &{Q, K, V, O, Up, Down, Gate}\\
LoRA Rank $r$          &16\\
LoRA Alpha $\alpha$        &32\\
\# Experts         &2\\
Routing strategy           &Soft Routing\\
\bottomrule
\label{Hyperparameter_OMoE}
\end{tabular}
\end{table}

\begin{table}[h]
\caption{Hyperparameter configurations of LoRA/DoRA and MixLoRA/MixDoRA. }
\label{table:hyperparameters_mixlora}
\centering
\begin{tabular}{lcc}
\toprule
\textbf{Hyperparameters}      & \textbf{LoRA/DoRA} & \textbf{MixLoRA/MixDoRA}   \\
\midrule
Cutoff Length &\multicolumn{2}{c}{512}\\
Learning Rate &\multicolumn{2}{c}{2e-4}\\
Optimizer     &\multicolumn{2}{c}{AdamW}\\
Batch size    &\multicolumn{2}{c}{16}\\
Accumulation Steps  &\multicolumn{2}{c}{8}\\
Dropout       &\multicolumn{2}{c}{0.05}\\
\# Epochs        &\multicolumn{2}{c}{2}\\
Where &\multicolumn{2}{c}{Q, K, V, O, Up, Down, Gate}\\
LoRA Rank $r$        &80  &16\\
LoRA Alpha $\alpha$       &160 &32\\
\# Experts        & - &8\\
Top-K         & -  &2\\
\bottomrule

\end{tabular}
\end{table}

\begin{table*}[ht!]
\centering
\caption{\label{appendix:singletask} An Overall comparison of different PEFT methods for single-task learning is provided. Most of the results are extracted from the original papers or reproduced by running the provided source code. The backbone model is LLaMA-2 7B. } 
{
\begin{tabular}{c|c|cccccc}
\hline
\textbf{Method}   &   \textbf{Params}  &      \textbf{ARC-e}   &     \textbf{ARC-c}   &   \textbf{BoolQ}   &   \textbf{OBQA}  &  \textbf{PIQA}    \\ 
\hline

\multicolumn{8}{c}{\textbf{\emph{Baselines}}}  \\
\hline


Parallel-Adapter  &    0.96\%   &      67.1    &    54.2    &     65.2    &   76.3    &   69.8     \\
Learned-Adapter   &   0.94\%   &     69.3     &    54.4  &    64.9    &  78.4   &    75.6     \\
P-tuning v2    &    0.97\%    &      63.5   &     51.3    &  61.2    &  76.1    &  66.2    \\

IAPT    &    0.96\%    &      66.3    &     54.7    &  67.8    &  79.2    &    77.3         \\
BitFit &   1.00\%   &      65.9     &    54.1    &   66.4   &   77.2    &     76.6    \\
(IA)$^{3}$  &    0.90\%   &     68.1     &   54.6      &    67.2    &  78.1    &   75.4    \\
SSP &   0.93\%   &     71.6    
 &  57.6    &     69.6    &    79.5      &    79.7     \\

AdaLoRA   &  0.92\%   &     73.8    &    57.9   &    69.2   &    80.4   &    82.1    \\

MOELoRA &  1.00\%    &  
 76.8   &  60.2   &     72.0     &    81.1    &     82.7    &     \\
MiLoRA    &   0.93\%       &   77.8   &  61.2   &  72.8  &  81.7 & 83.3   \\

MiDoRA   &   0.93\%       &   77.5     &    61.3     &   72.9    &  81.3 & 83.1     \\

\hline
\multicolumn{8}{c}{\textbf{\emph{Our proposed methods}}}  \\
\hline

OMoE-LoRA (ours)   &   0.73\%    &     \underline{79.3}     &   56.6   &  \textbf{73.5}   &  80.6  &  \textbf{84.5}   \\

OMoE-DoRA (ours)   &   0.73\%    &     \textbf{80.1}     &   55.3     &    \underline{73.1}     &   80.0    &     \underline{83.3}       \\

\hline
\end{tabular}}
\end{table*}

\textbf{Computing infrastructure} We run all our experiments on NVIDIA A6000 (48GB) GPUs and set up with Python 3.10 and Ubuntu 20.04 on x86-64 CPUs. 

\textbf{Pretrained backbones} The main experiments use the most recent open-sourced LLMs, LLaMA-2 7B as the pretrained backbone model. In the ablation studies, we will also use the recently released LLaMA-2 8B and LLaMA-2 13B. Prediction heads When fine-tuning LLaMA-2 7B, we only consider the supervised fine-tuning setting.

\textbf{Hyper-parameters for OMoE} In our experiments, unless otherwise specified, we set hyperparameters as illustrated in Table~\ref{Hyperparameter_OMoE}. Under the above setting, OMoE will introduce 0.73\% tunable parameters to the LLaMA-2 7B backbone.

\textbf{Hyper-parameters for LoRA/DoRA and MixLoRA/MixDoRA} In our experiments, we set hyperparameters for LoRA/DoRA and MixLoRA/MixDoRA as illustrated in Table~\ref{table:hyperparameters_mixlora}, consistent with the settings in~\cite{li2024mixlora}.

\textbf{Hyper-parameters for other baselines} In our experiments, we set hyperparameter for other baselines as follows, consistent with the settings in~\cite{zhang2024milora}:
\begin{itemize}
\item P-tuning V2: The number of prompt tokens at each layer is set to 16, and the soft prompts are initialized with dimension 640, and then are projected to dimension 4096.
\item IAPT: The prompt length is 4, and the bottleneck dimension for the prompt generator is 320.
\item Parallel-Adapter and Learned-Adapter: The bottleneck dimension is set to 160. Adapters are connected to both the self-attention and FFN sub-layer.
\item SSP: Simply adjust the sparsity for SSP to reduce parameters.
\item BitFit: The bias vectors are initialized with dimension 64, and then a learnable projection layer projects it to the same dimension with the LLama-2 backbone.
\item $\text{IA}^3$: The activation adjusting vectors are added the Query, Key, and Up activations. The adjusting vectors are initialized with dimension 128, and then a learnable projection layer projects it to the same dimension with the Llama-2 backbone.
\item AdaLoRA: The initial rank at each module is set to 64, and half of the rank budget is pruned during fine-tuning.
\item MoELoRA: The rank size r at each LoRA module is set to 32.
\end{itemize}

 \section{Additional Results on Other Baselines}
 \label{Results_other_baselines}
In the main paper, we compared OMoE with three widely recognized and well-performing baselines: LoRA, DoRA, and MixLoRA, across the LLaMA-2 7B, 8B, and 13B models. Here, in addition to the three baselines presented in Table~\ref{table:single_task_results}, we provide results from experiments involving 11 additional strong baselines on the Llama 7B backbone in Table~\ref{appendix:singletask}. The results indicate that OMoE demonstrates significant improvements in both parameter efficiency and performance compared to the baselines.

\section{Additional Visualization of Representations using Different Methods}
\label{differentmethod}
\begin{figure*}[]  
\centering
\subfigure[Soft Routing]{
\begin{minipage}[b]{0.3\textwidth}
\includegraphics[width=1\textwidth]{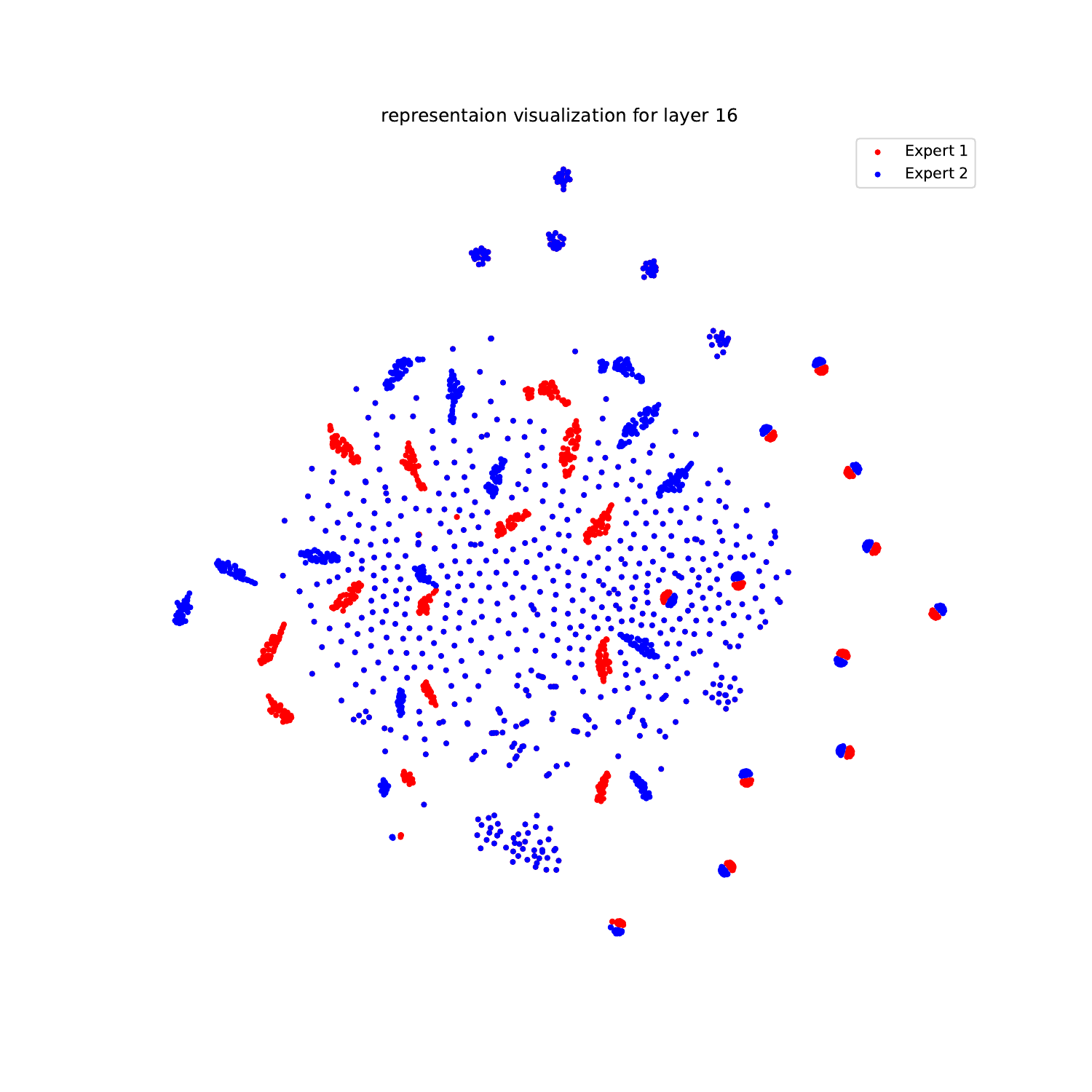} 
\end{minipage}
}
\subfigure[Top-2 Routing]{
\begin{minipage}[b]{0.3\textwidth}
\includegraphics[width=1\textwidth]{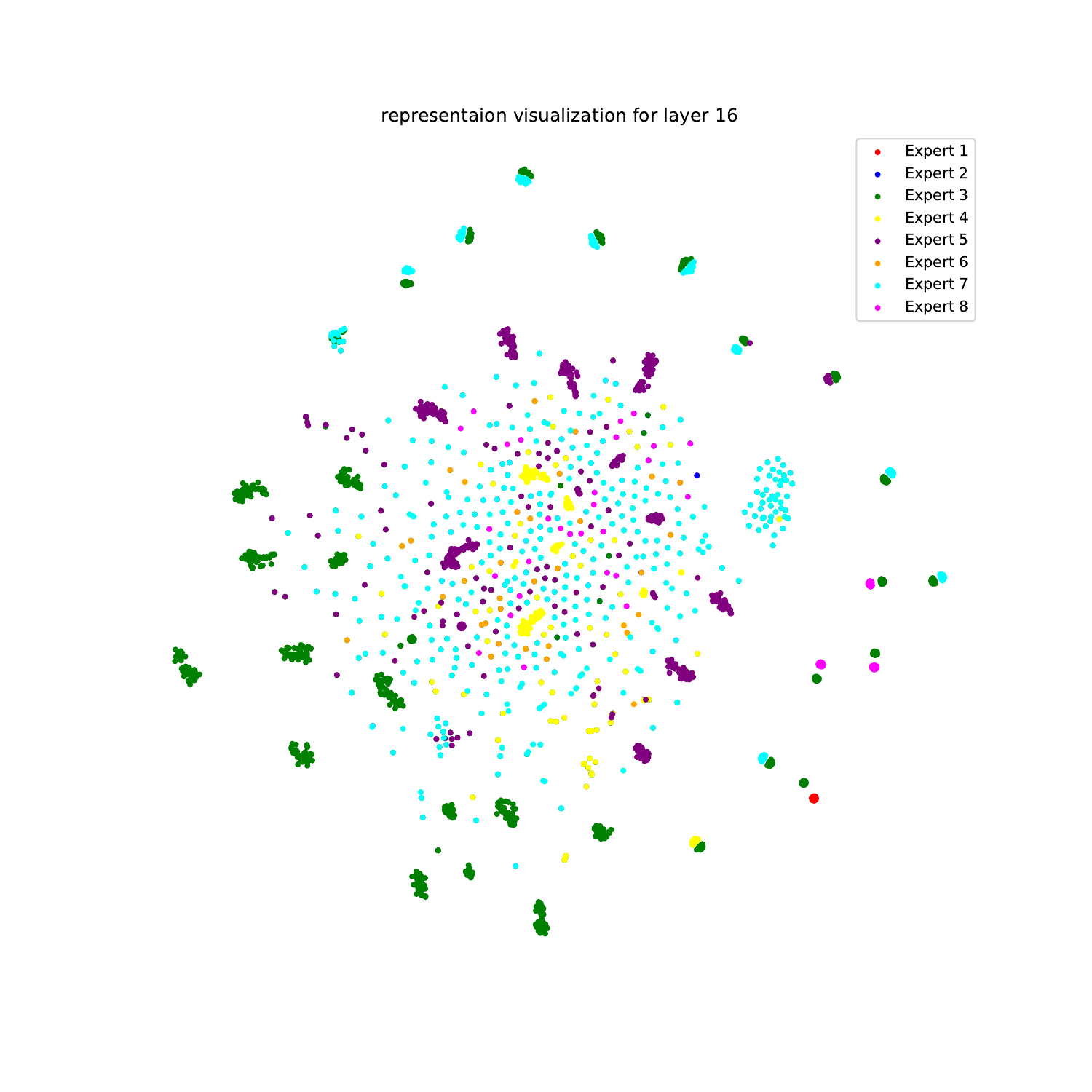} 
\end{minipage}
}
\subfigure[OMoE ]{
\begin{minipage}[b]{0.3\textwidth}
\includegraphics[width=1\textwidth]{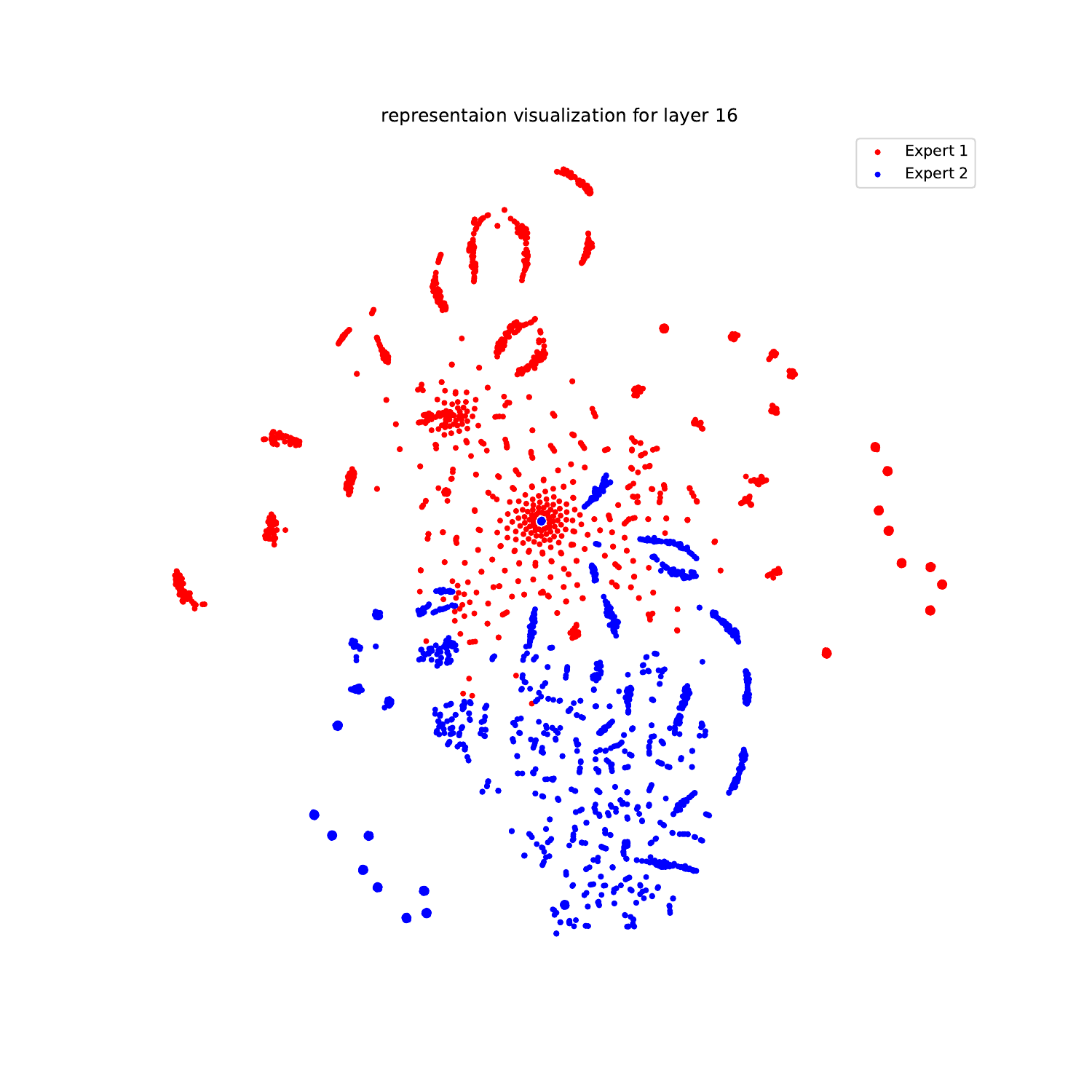} 
\end{minipage}
}
\caption{The representations of two activated experts using soft routing, Top-2 routing and OMoE. The representations are obtained from layer 16 of LLaMA-2 7B.}
\label{fig:layer16}
\end{figure*}

\begin{figure*}[]  
\centering
\subfigure[Soft Routing]{
\begin{minipage}[b]{0.3\textwidth}
\includegraphics[width=1\textwidth]{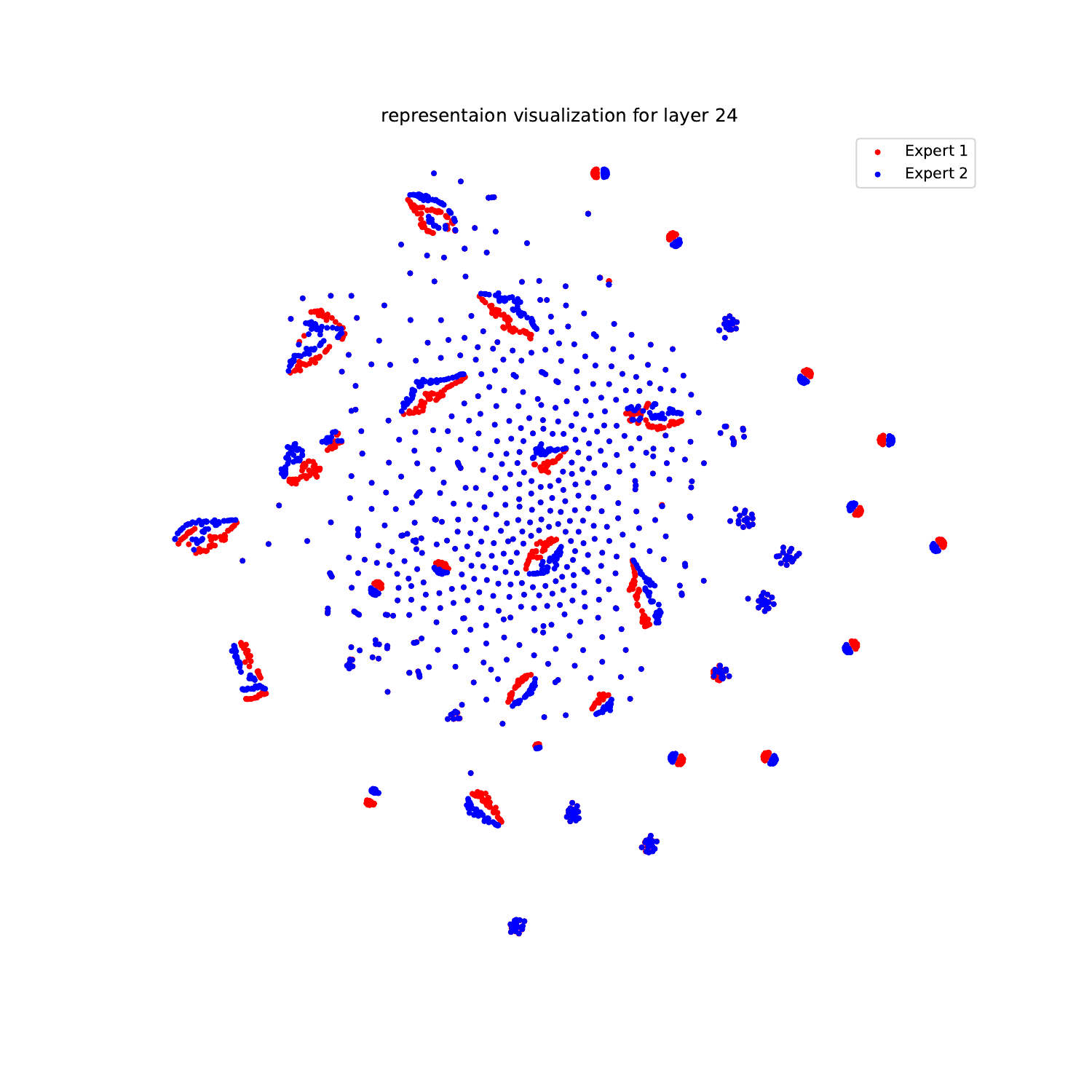} 
\end{minipage}
}
\subfigure[Top-2 Routing]{
\begin{minipage}[b]{0.3\textwidth}
\includegraphics[width=1\textwidth]{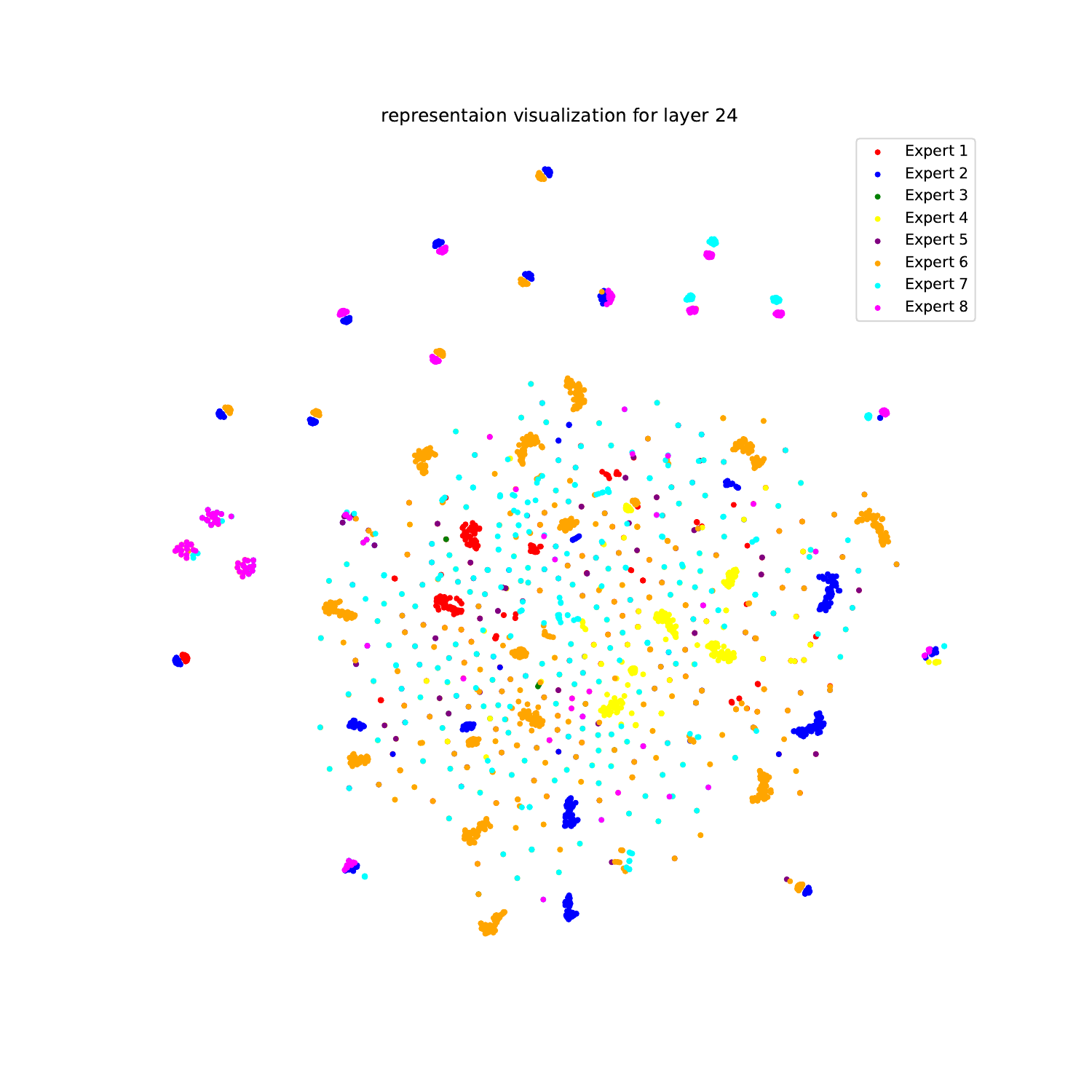} 
\end{minipage}
}
\subfigure[OMoE ]{
\begin{minipage}[b]{0.3\textwidth}
\includegraphics[width=1\textwidth]{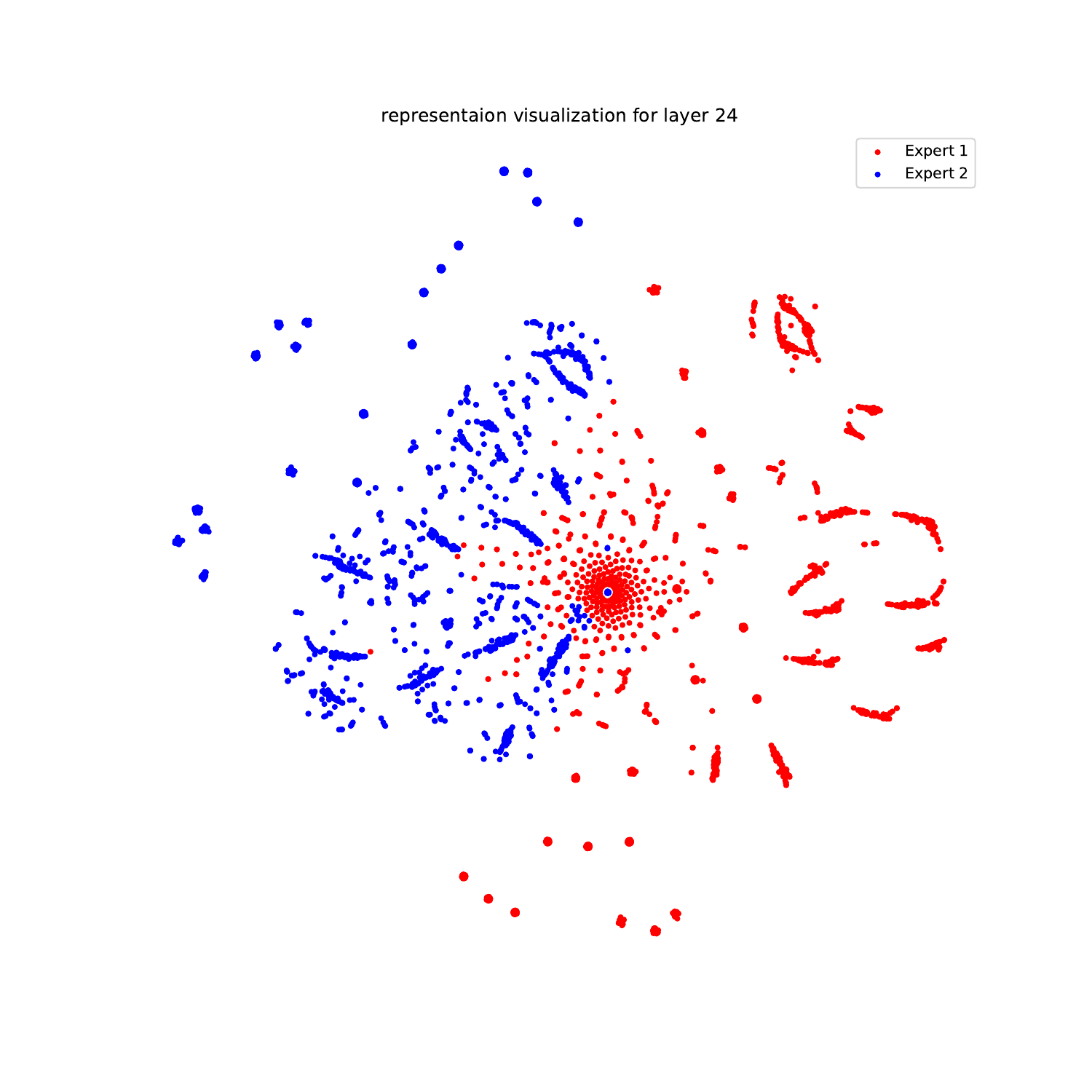} 
\end{minipage}
}
\caption{The representations of two activated experts using soft routing, Top-2 routing and OMoE. The representations are obtained from layer 24 of LLaMA-2 7B.}
\label{fig:layer24}
\end{figure*}

\begin{figure*}[]  
\centering
\subfigure[Soft Routing]{
\begin{minipage}[b]{0.3\textwidth}
\includegraphics[width=1\textwidth]{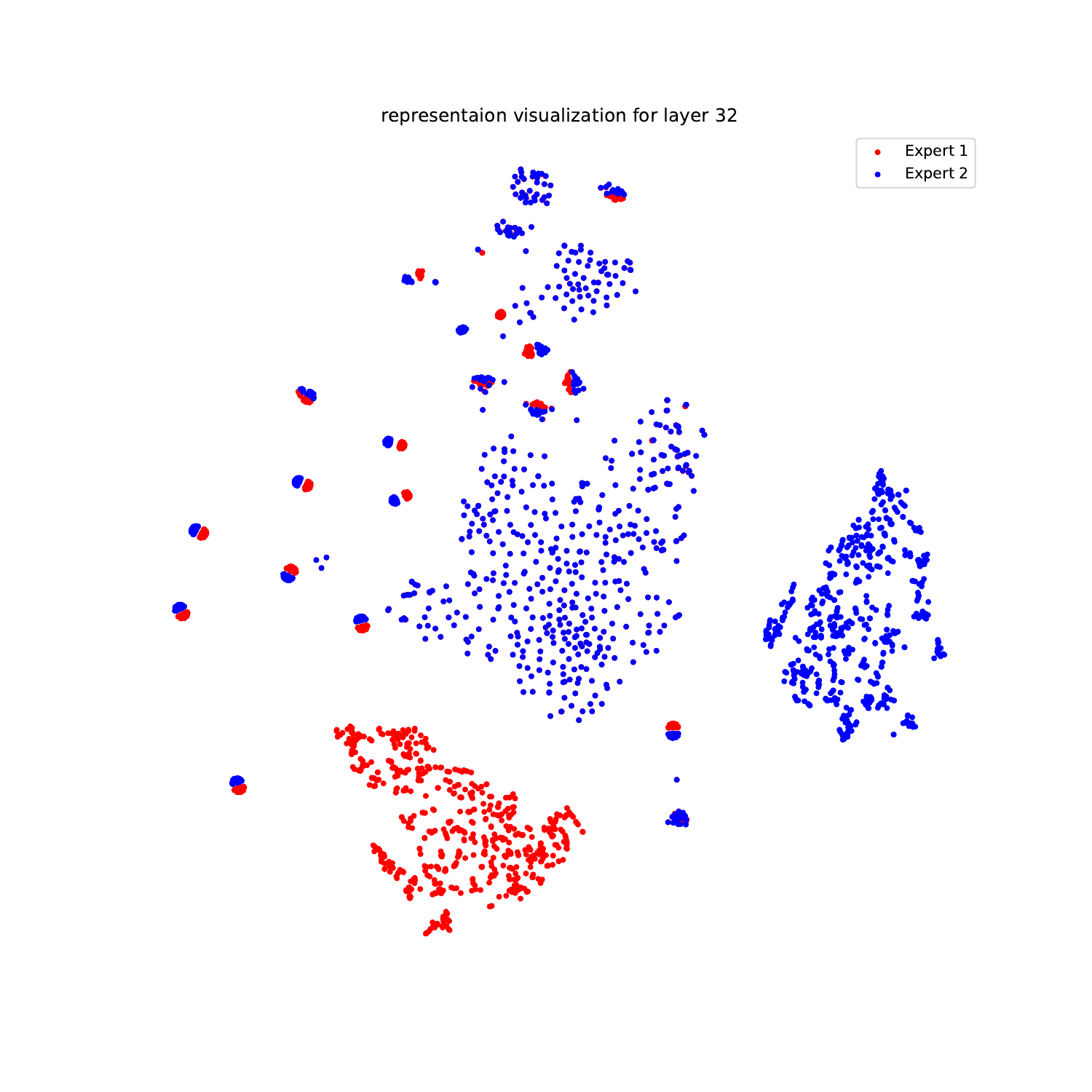} 
\end{minipage}
}
\subfigure[Top-2 Routing]{
\begin{minipage}[b]{0.3\textwidth}
\includegraphics[width=1\textwidth]{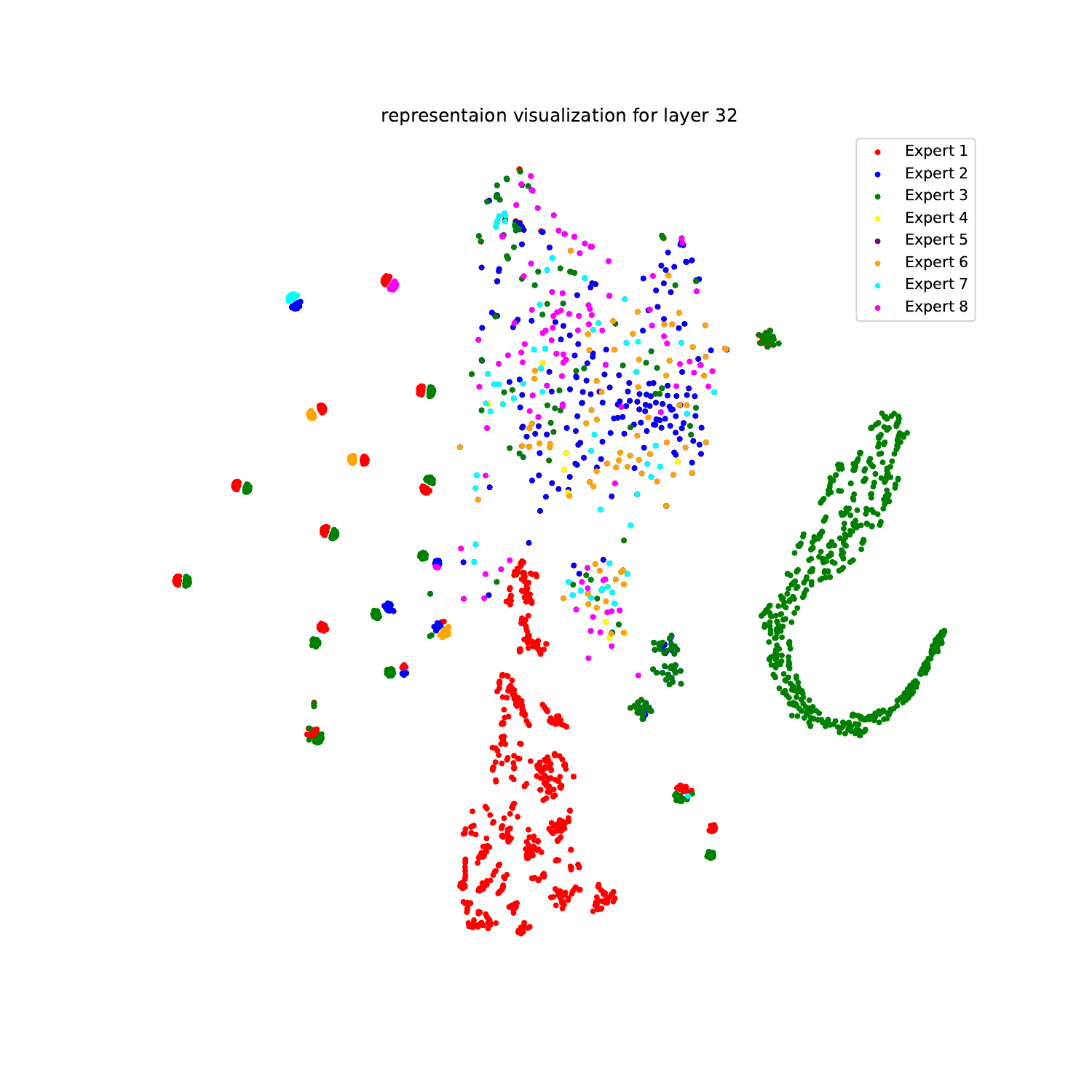} 
\end{minipage}
}
\subfigure[OMoE ]{
\begin{minipage}[b]{0.3\textwidth}
\includegraphics[width=1\textwidth]{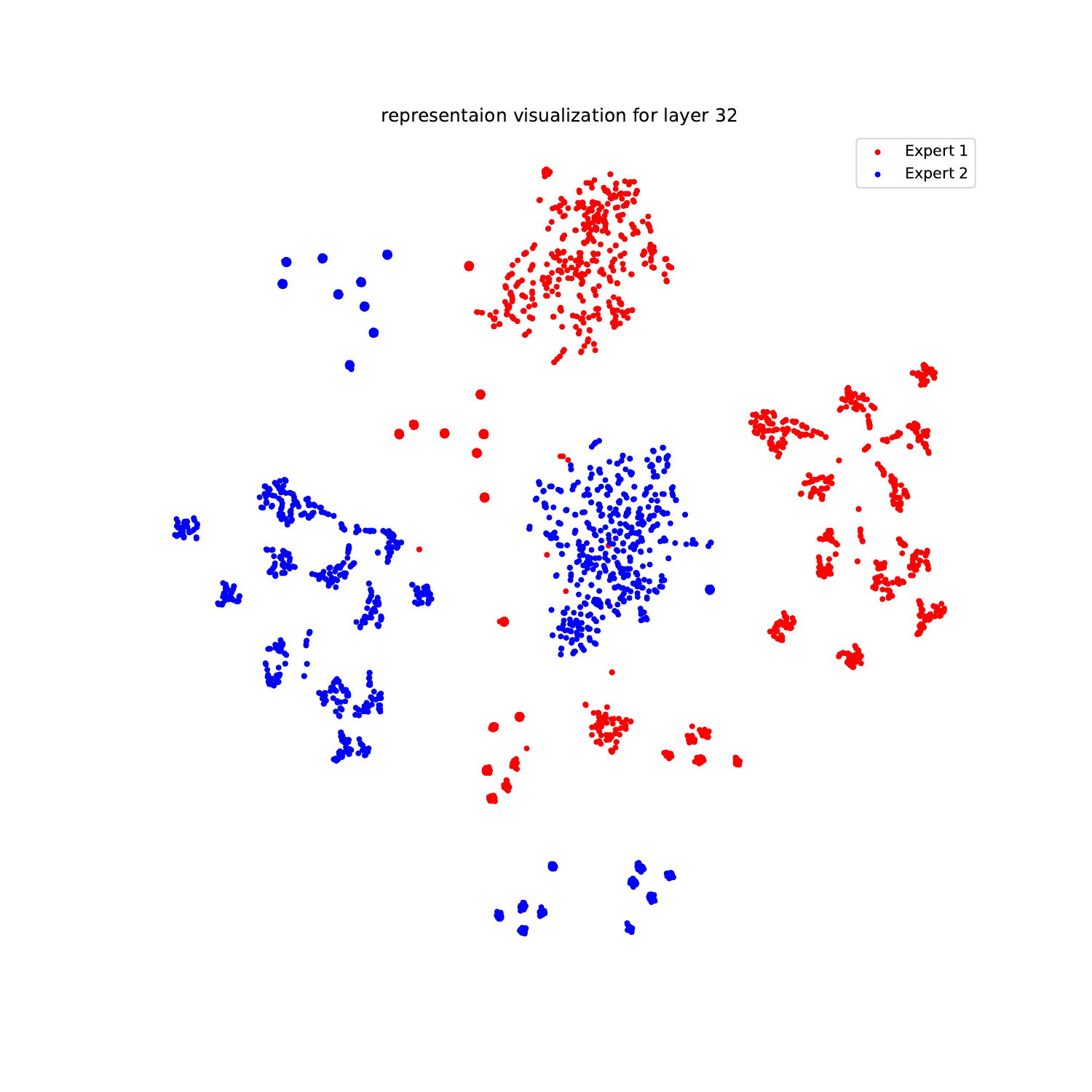} 
\end{minipage}
}
\caption{The representations of two activated experts using soft routing, Top-2 routing and OMoE. The representations are obtained from layer 24 of LLaMA-2 7B.}
\label{fig:layer32}
\end{figure*}

We visualize representations of two activated experts using soft routing, Top-2 routing and our OMoE model, while Top-2 routing utilizes a total of 8 experts. As illustrated in Fig.~\ref{fig:layer16}, Fig.~\ref{fig:layer24} and Fig.~\ref{fig:layer32}, the representations are obtained from layer 16, 24, 32 of LLaMA-2 7B by projecting the vectors into 2D space via t-SNE for qualitative analysis. 

\section{Additional Visualization of Representations using Different Number of Experts in OMoE}
\label{differenk}
\begin{figure*}[]  
\centering
\subfigure[2 experts]{
\begin{minipage}[b]{0.3\textwidth}
\includegraphics[width=1\textwidth]{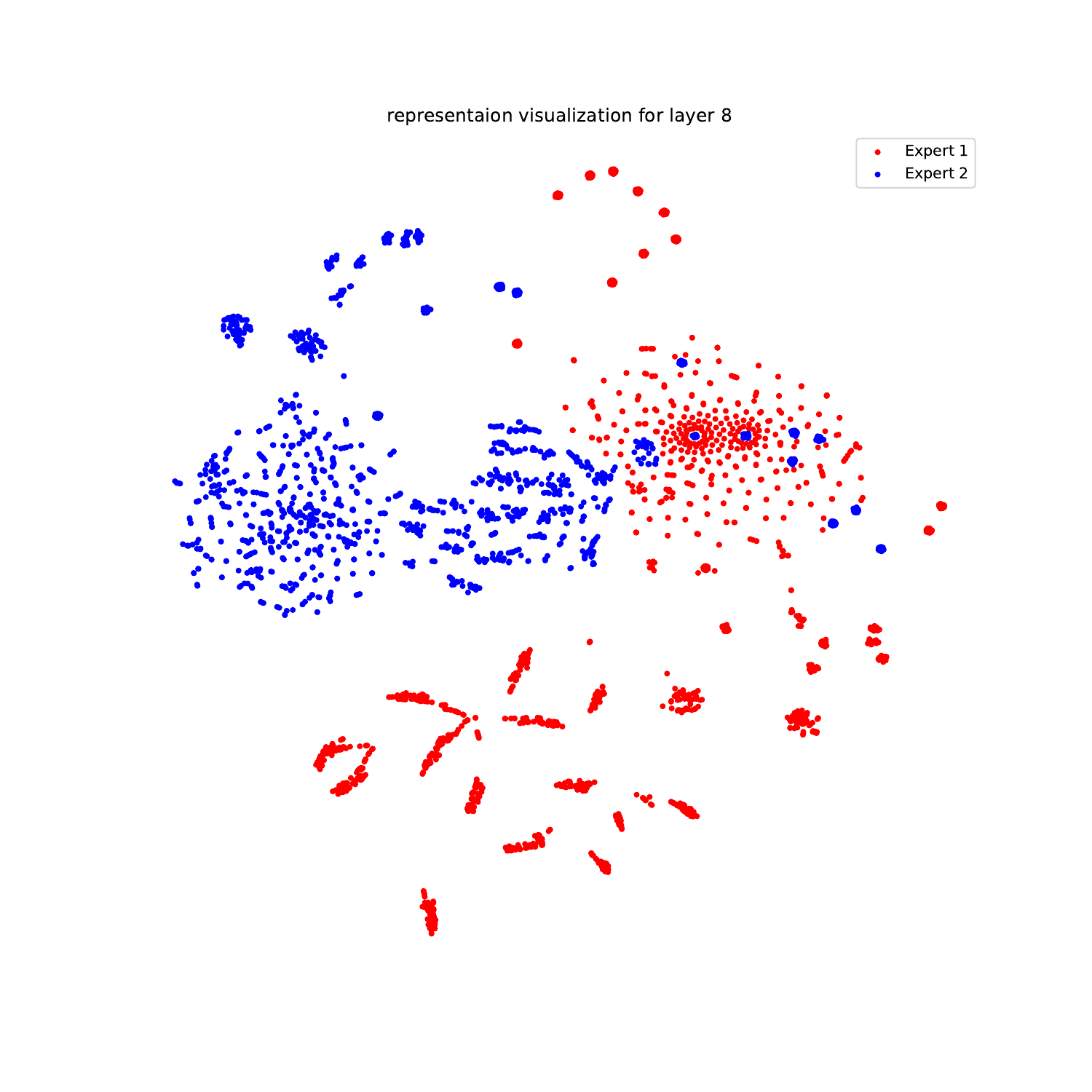} 
\end{minipage}
}
\subfigure[3 experts]{
\begin{minipage}[b]{0.3\textwidth}
\includegraphics[width=1\textwidth]{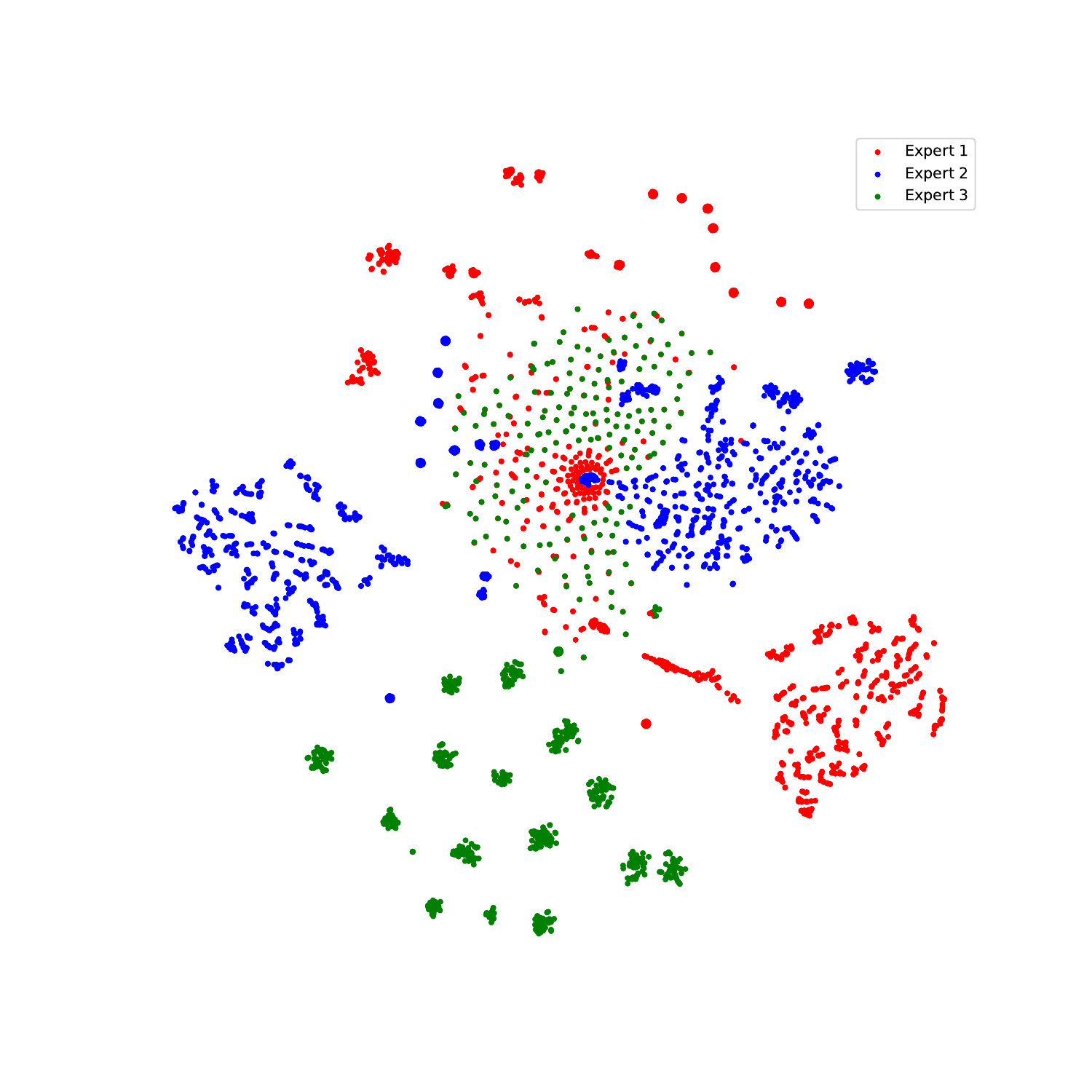} 
\end{minipage}
}
\subfigure[4 experts]{
\begin{minipage}[b]{0.3\textwidth}
\includegraphics[width=1\textwidth]{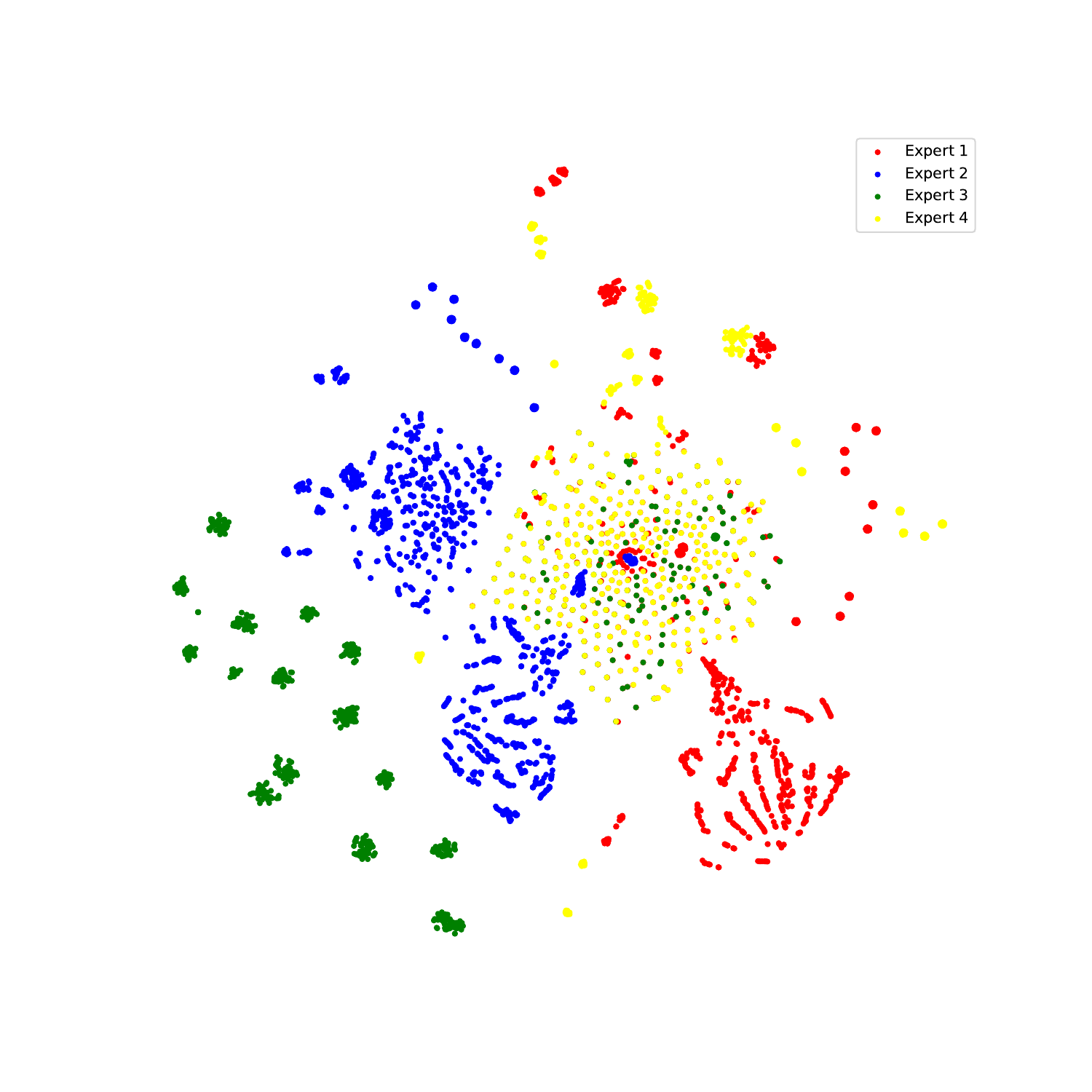} 
\end{minipage}
}
\caption{The representations under different numbers of experts in OMoE using top-2 routing are visualized via t-SNE. The representations are obtained from layer 8 of LLaMA-2 7B.}
\label{layer8_Omoe-expert}
\end{figure*}

\begin{figure*}[]  
\centering
\subfigure[2 experts]{
\begin{minipage}[b]{0.3\textwidth}
\includegraphics[width=1\textwidth]{appendix_images/3_orth_test.pdf} 
\end{minipage}
}
\subfigure[3 experts]{
\begin{minipage}[b]{0.3\textwidth}
\includegraphics[width=1\textwidth]{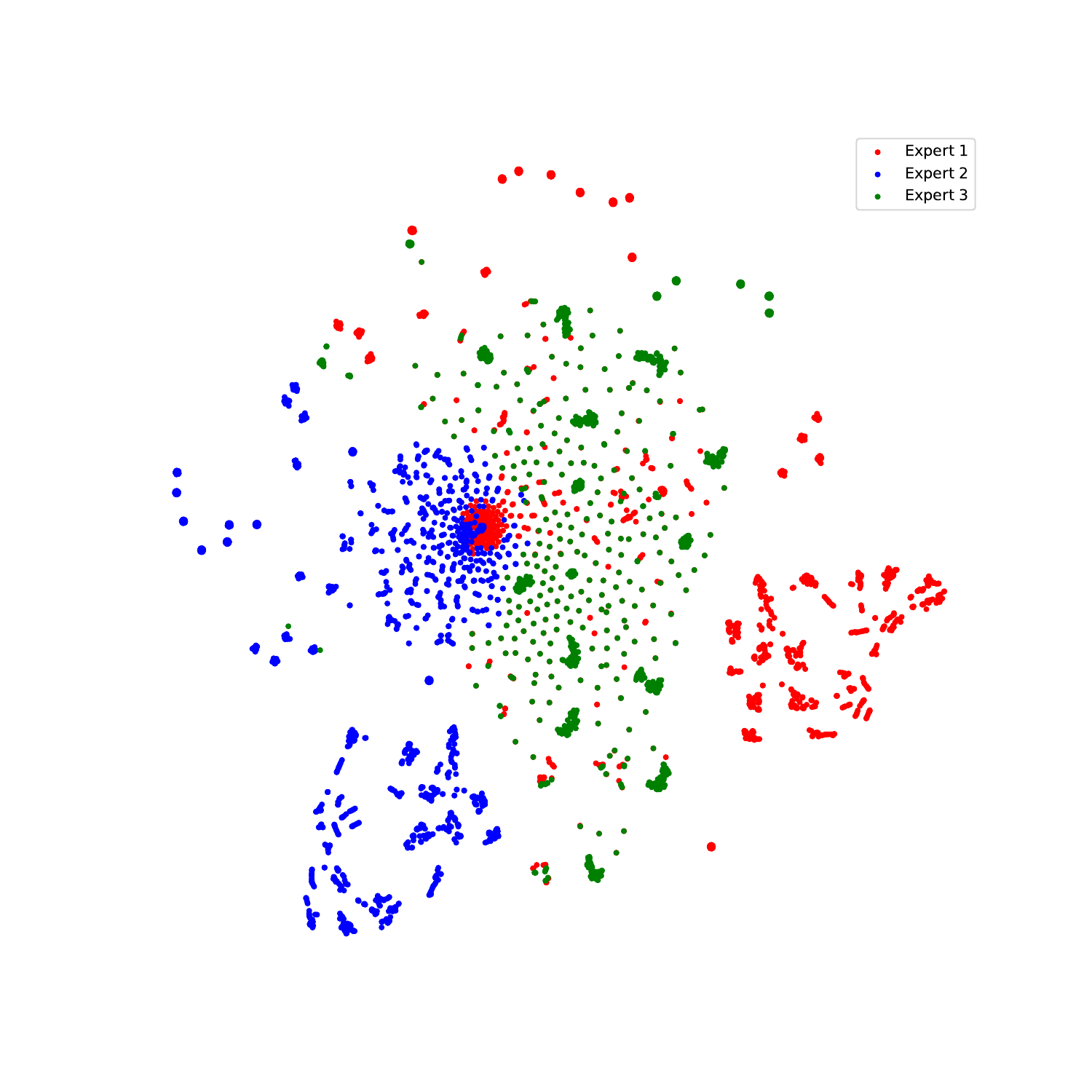} 
\end{minipage}
}
\subfigure[4 experts]{
\begin{minipage}[b]{0.3\textwidth}
\includegraphics[width=1\textwidth]{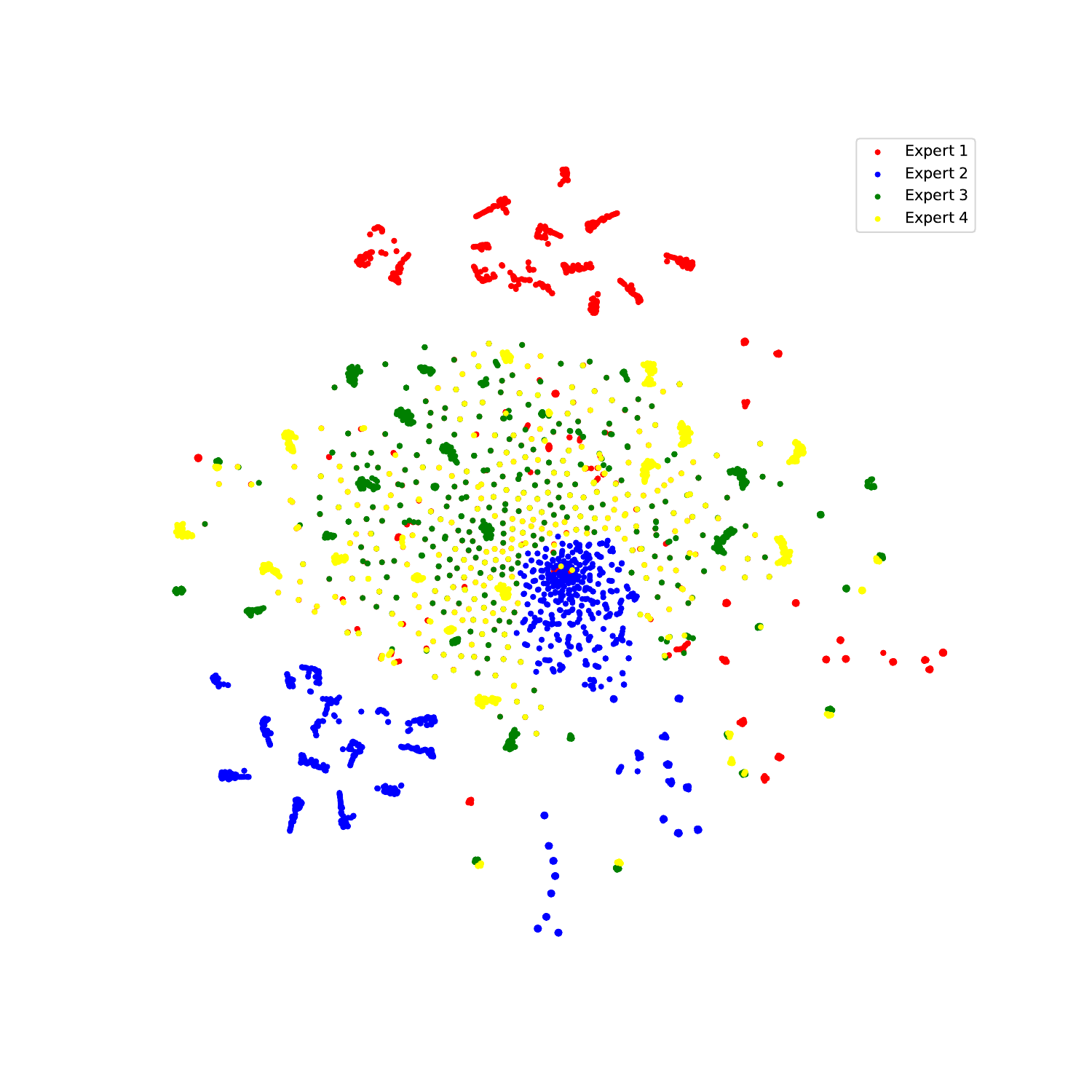} 
\end{minipage}
}
\caption{The representations under different numbers of experts in OMoE using top-2 routing are visualized via t-SNE. The representations are obtained from layer 24 of LLaMA-2 7B.}
\label{layer24_Omoe-expert}
\end{figure*}

\begin{figure*}[ht]  
\centering
\subfigure[2 experts]{
\begin{minipage}[b]{0.3\textwidth}
\includegraphics[width=1\textwidth]{appendix_images/4_orth_test.pdf} 
\end{minipage}
}
\subfigure[3 experts]{
\begin{minipage}[b]{0.3\textwidth}
\includegraphics[width=1\textwidth]{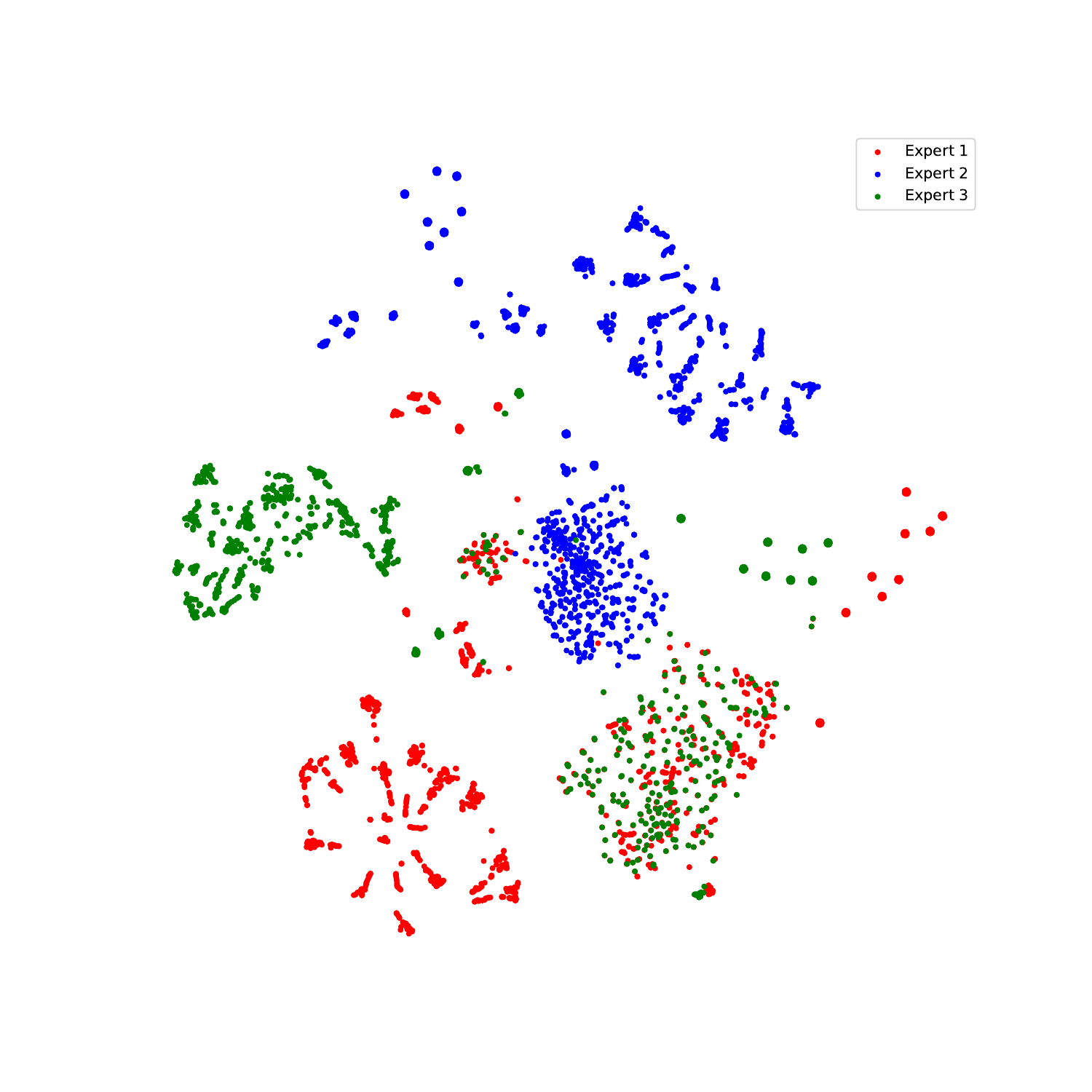} 
\end{minipage}
}
\subfigure[4 experts]{
\begin{minipage}[b]{0.3\textwidth}
\includegraphics[width=1\textwidth]{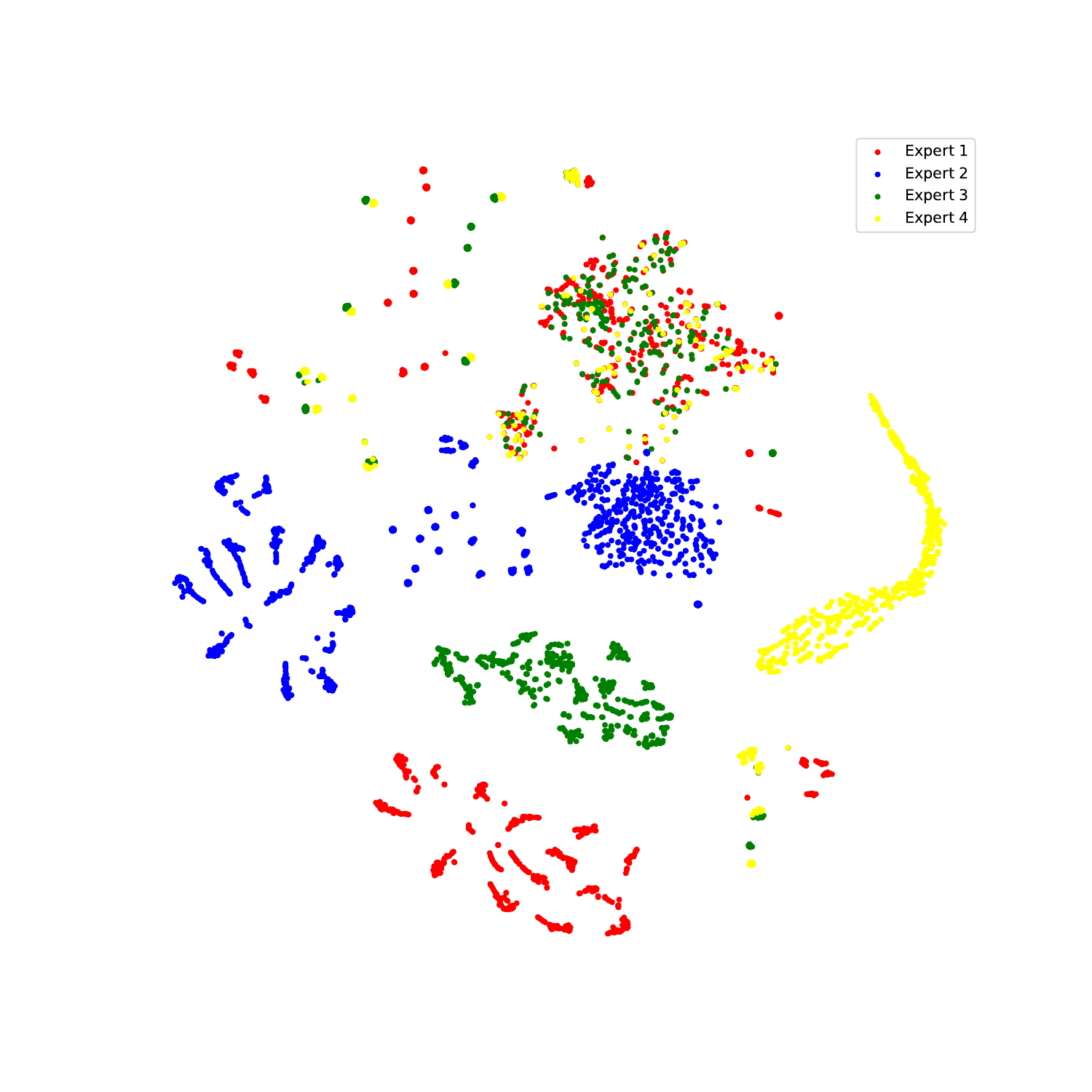} 
\end{minipage}
}
\caption{The representations under different numbers of experts in OMoE using top-2 routing are visualized via t-SNE. The representations are obtained from layer 32 of LLaMA-2 7B.}
\label{layer32_Omoe-expert}
\end{figure*}

As illustrated in Fig.~\ref{layer8_Omoe-expert}, Fig.~\ref{layer24_Omoe-expert} and Fig.~\ref{layer32_Omoe-expert}, the representations under different numbers of experts in OMoE are visualized using a uniform t-SNE transformation for 2D representation reduction. The representations are obtained from layers 8, 24, and 32 of LLaMA-2 7B.


\end{document}